\documentclass[11pt,a4paper,table,dvipsnames]{article}
\usepackage{acl}
\usepackage{times,latexsym}
\usepackage{url}
\usepackage[T1]{fontenc}
\usepackage{booktabs}
\usepackage{tabularx}
\usepackage{xcolor}
\usepackage{array}
\usepackage{amsmath}
\usepackage{bbm}
\usepackage{graphicx}
\usepackage{caption}
\usepackage{subcaption}
\usepackage{siunitx}
\usepackage{multirow}
\usepackage{url}
\usepackage{microtype}
\usepackage{makecell}
\usepackage{xspace}
\usepackage{xltabular}
\usepackage{amssymb}
\usepackage{pifont}
\usepackage{scalerel}
\usepackage{xparse}
\usepackage[english]{babel}

\newcommand{\hatecheck}{\textsc{HateCheck}\xspace}
\newcommand{\checklist}{\textsc{CheckList}\xspace}

\newcommand{\D}{\ensuremath{\mathcal{D}}\xspace}
\newcommand{\g}[1][]{\ensuremath{\text{G}_{\text{#1}}}\xspace}
\newcommand{\fscore}{\ensuremath{F_1}\xspace}
\newcommand{\gfunc}{\g[func]}
\newcommand{\gclass}{\g[class]}

\newcommand{\gseen}{\g[seen]}

\newcommand{\T}{\ensuremath{\mathcal{T}}\xspace}
\newcommand{\nfunc}{\ensuremath{n_\text{func}}\xspace}
\newcommand{\nclass}{\ensuremath{n_\text{class}}\xspace}
\newcommand{\F}{\ensuremath{\mathcal{F}}\xspace}
\newcommand{\C}{\ensuremath{\mathcal{C}}\xspace}
\newcommand{\cmark}{\ding{51}}%
\newcommand{\xmark}{\ding{55}}%
\newcommand{\sbase}{\ensuremath{s_\text{Base}}\xspace}
\newcommand{\sseen}{\ensuremath{s_\text{Seen}}\xspace}
\newcommand{\sfunc}{\ensuremath{s_\text{Func}}\xspace}
\newcommand{\sclass}{\ensuremath{s_\text{Class}}\xspace}
\NewDocumentCommand\hface{}{\scalerel*{\includegraphics{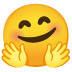}}{X}}
\NewDocumentCommand\bflag{}{\scalerel*{\includegraphics{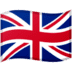}}{X}}
\NewDocumentCommand\heels{}{\scalerel*{\includegraphics{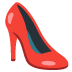}}{X}}

\title{Functionality learning through specification instructions}

\author{Pedro Henrique Luz de Araujo$^\diamond$ 
\and
  Benjamin Roth$^\dagger$
  \\
  \ \\
  $^\diamond$$^\dagger$Faculty of Computer Science, University of Vienna, Vienna, Austria
  \\
  $^\diamond$UniVie Doctoral School Computer Science, Vienna, Austria
  \\
  $^\dagger$Faculty of Philological and Cultural Studies, University of Vienna, Vienna, Austria
  \\
  \texttt{\{pedro.henrique.luz.de.araujo, benjamin.roth\}@univie.ac.at} \\ 
}

\date{}

\begin{document}
\maketitle
\begin{abstract}
  Test suites assess natural language processing models' performance on specific functionalities: cases of interest involving model robustness, fairness, or particular linguistic capabilities.
  This paper introduces specification instructions: text descriptions specifying fine-grained task-specific behaviors.
  For each functionality in a suite, we generate an instruction that describes it.
  We combine the specification instructions to create specification-augmented prompts, which we feed to language models pre-trained on natural instruction data.

We conduct experiments to measure how optimizing for some functionalities may negatively impact functionalities that are not covered by the specification set.
Our analyses across four tasks and models of diverse sizes and families show that smaller models struggle to follow specification instructions.
However, larger models (>~3B params.) can benefit from specifications and---surprisingly---even generalize certain desirable behaviors across functionalities.\footnote{Our code is available on \url{https://github.com/peluz/specification-instructions}}
\end{abstract}

\begin{figure}[tb]
  \centering
  \includegraphics[width=\linewidth]{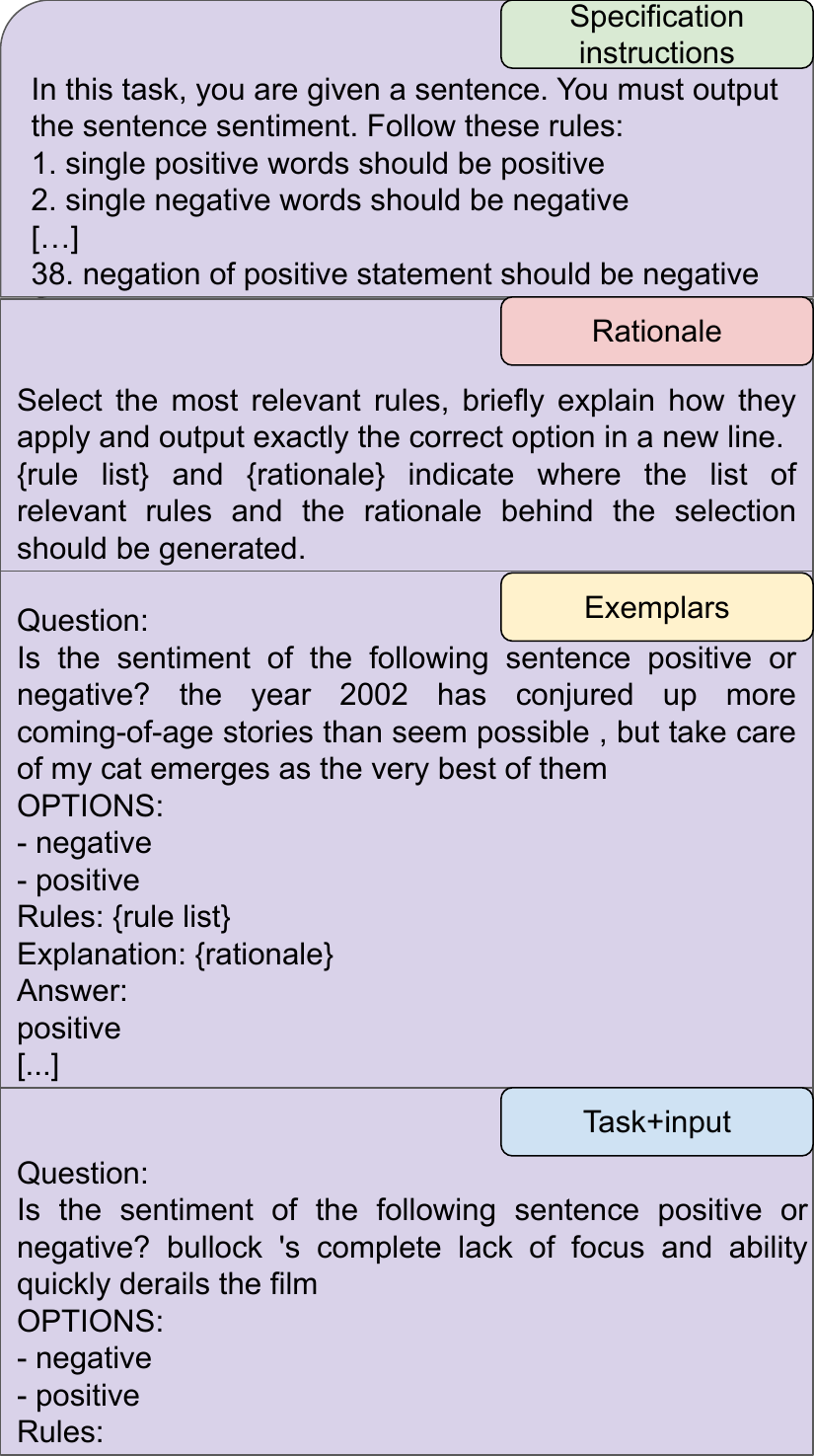}
  \caption{Example of a specification-augmented prompt for sentiment analysis. Each module adds information about how the task is expected to be performed.}
  \label{fig:promptExample}
\end{figure}

\section{Introduction}
Test suites \cite{kirk2022hatemoji,rottger2021hatecheck,ribeiro2020accuracy,mccoy2019right} have been proposed as an evaluation framework to test for specific functionalities in natural language processing (NLP) models.
Each functionality is a set of test cases, generally input-output pairs, relating to a particular aspect of a task.
For example, a test suite for hate speech detection can assess distinct expressions of hate (e.g., implicit derogation), while a sentiment analysis suite can measure how well a model handles specific phenomena (e.g., negation).
Suites complement the standard practice of evaluating on a held-out test set from the same distribution of the training set \cite{linzen2020accelerate}.
If the latter is representative of the underlying task distribution, it is a good measure of average correctness; suites, on the other hand, allow for in-depth evaluation of relevant phenomena that may be underrepresented in general data.

Though test suites can point to failure cases, there are no clear guidelines on how to act upon their feedback to develop more robust and trustworthy models.
Data augmentation has been suggested as a potential avenue for improvement \cite{rottger2021hatecheck} by including additional training cases that correspond to the suite's cases.
However, constructing or annotating instances targeting specific functionalities is costly, and further training models is expensive for large models and infeasible for closed-source ones.
Furthermore, fine-tuning models on suites' test cases has been shown to help seen functionalities \cite{ribeiro2022adaptive,malon2022fast,liu2019inoculation}, but often does not generalize to unseen ones and harms general performance \cite{luzdearaujo2023cross,luzdearaujo2022checking,rozen2019diversify,mccoy2019right}. 

Missing from the literature are analyses for the increasingly influential paradigm of prompting large language models (LLMs) \cite{liu2023pretrain}, which has superior zero- and few-shot capabilities, particularly for models trained on natural language instructions \cite{ouyang2022training}.
Prompting with functionality information may improve relevant aspects of model behavior with no need for fine-tuning, which requires additional training data and computational resources.
Since the model parameters are not updated, prompting is also less vulnerable to overfitting to seen functionalities, a substantial limitation in previous work.

This paper explores specification instructions and their effect on functionality performance.
Contrary to previous efforts, we do not expose the model to suite examples or fine-tune it.
Instead, we elicit the desired behavior by augmenting prompts with instructions that specify the suite's functionalities.
For example, if a sentiment analysis suite contains a functionality that tests whether predictions are invariant to nationalities mentioned in the input, an instruction such as ``nationality should be irrelevant to sentence sentiment'' would be added to the task prompts.

Our main contributions are:

\textbf{1.}~Creating two sets---handcrafted and machine-generated---of 144 specification instructions for 4 test suites from different tasks (sentiment analysis, paraphrase identification, reading comprehension, and hate speech detection\footnote{\textbf{This paper contains examples
of abusive and hateful language.}}) and designing specification-augmented prompts.

\textbf{2.}~Assessing the impact of the specification-augmented prompts for seven models ranging from 80M to billions of parameters and covering three model families.

\textbf{3.}~Evaluating cross-functionality impact through scenarios with held-out specifications, finding that overfitting to seen cases is much less of a concern here than in the fine-tuning paradigm.

\textbf{4.}~Qualitatively examining the impact of specification-augmented prompts and the interplay between different specifications by examining which functionalities are most helped or harmed across different evaluation scenarios.


\section{Prompting with specification instructions}
\subsection{Problem setting}
We consider a task to be composed of a dataset $\D$ of $n$ labeled examples assumed to be identically and independently distributed (i.i.d.), $\D = \{(x_i, y_i)\}_{i=1}^n$, and a test suite $\T$ of $m$ test cases $\{t_i\}_{i=1}^m$ partitioned into $\nfunc$ disjoint functionalities $\{\F_i\}_{i=1}^{\nfunc}$. Each functionality is assigned to a functionality class $c \in \{\C_i\}_{i=1}^{\nclass}$, such that $\nclass < \nfunc < m$.
While $\D$ describes the general behavior expected for the task, $\T$ specifies fine-grained aspects of the expected behavior.

For example, $\D$ can be a dataset of tweets with labels indicating whether they contain hate speech; $\T$ would be a suite with functionalities that assess specific expressions of hate (e.g., use of profanity, threatening language) and contrastive non-hate speech (e.g., use of reclaimed slurs, non-hateful profanity) \cite{rottger2021hatecheck}.

The functionality classes encompass coarse-grained dimensions such as fairness and robustness, the functionalities assess finer-grained aspects such as gender fairness and robustness to typos, and the test cases operationalize them as pairs of inputs and expected behaviors~\cite{rottger2021hatecheck,ribeiro2020accuracy}.


\subsection{Prompt modules}
In our setting, each prompt is composed of several modules: a necessary task description and optional modules that further specify the task.
More formally, given an input $x$, a task description $\tau$, and a (possibly empty) set of optional modules $M$, we have
\begin{equation}
  z = f(x, \tau, M)\,,
\end{equation} 
where $z$ is the resulting prompt, and $f$ is a function that combines prompt modules and inputs.
$x$ and $\tau$ are strings, and $M$ is a set of strings.
Fig.~\ref{fig:promptExample} shows a prompt for sentiment analysis with all optional modules.

Below, we describe each prompt module:

\textbf{Task description (Task)}:
a natural language instruction that describes the task. For example, ``Answer a question about this article:'' for a reading comprehension task (details in Appendix~\ref{sec:promptModules}).

\textbf{Exemplars (Ex)}:
input-label pairs that exemplify the task, also known as demonstrations \cite{brown2020language}.
They can help to improve task performance by providing the model with information about the task format, label space, input distribution, and input-label mapping ~\cite{min2022rethinking}.

\textbf{Specification instructions (Spec)}:
provides, for each functionality in the suite, an instruction that specifies the behavior expected by the functionality (e.g., ``typos in the question are irrelevant to the answer'').\footnote{We discuss specification instruction generation in §\ref{sec:ruleGen} and show all specification instructions we generate in Appendix \ref{sec:rules}.}
Their purpose is to elicit the LM to generate text that conforms to what the suite specifies.

\textbf{Rationales (Rat)}:
asks the model to state the applicable specifications and the underlying rationale (before generating the task prediction).
This module is similar to chain-of-thought prompting~\cite{kojima2022Large,wei2022Chain}, which asks the LM to work the solution to a problem step-by-step and has been shown to improve LM performance in reasoning benchmarks.

Combinations of optional modules yield different \textit{prompting methods}.
We explore two baseline methods with no suite information (\textbf{Task}, \textbf{Task+Ex}) and four specification-augmented methods (\textbf{Task+Spec}, \textbf{Task+Spec+Ex}, \textbf{Task+Spec+Rat}, \textbf{Task+Spec+Rat+Ex}).
By comparing the baselines and specification-augmented methods, we assess the impact of incorporating additional task specifications; by comparing the specification-augmented methods, we investigate the impact of the individual modules.

\subsection{Cross-functional analysis}
\label{sec:crossFunc}
The specification instructions do not cover all aspects of desired task behavior---there is always a chance that important phenomena are (intentionally or not) left unspecified.\footnote{If one could completely specify a task, training a model would be unnecessary.} 
For example, the specifications for sentiment analysis (§~\ref{sec:tasks}) state that sentence sentiment is invariant to persons' names and locations.
However, 
sentiment should also be invariant to organization names (not checked in the specifications).
An evaluation setting that measures performance only on included specifications cannot examine how the instructions affect specifications the model developer did not think to include.

To address this, we adapt the cross-functional analysis method \cite{luzdearaujo2023cross} to the prompt-based learning paradigm.
The method was originally proposed for the fine-tuning learning paradigm and involves training and evaluating on different sets of functionalities.
To translate this into the prompting paradigm, we vary which specifications are included in each prompt:

\textbf{Seen scenario}: No specifications are held-out. 
This scenario measures how including specifications affects performance for \textit{seen} functionalities.

\textbf{Functionality generalization}: We remove the specification that applies to the input.
For example, if the input belongs to Functionality 1, we remove specification instruction 1 from the prompt.
This scenario estimates performance for \textit{unseen} functionalities.

\textbf{Functionality class generalization}: We remove all specifications from the same functionality class of the applicable specification.
In the example above, if functionalities 1 to 3 are from the same functionality class, we remove specifications 1 to 3 from the prompt.
This scenario estimates performance for \textit{unseen} functionality classes.

\section{Experimental setting}

\begin{table*}[tb]
    \footnotesize
    \centering
    \rowcolors{2}{gray!20}{white}
    \begin{tabularx}{\linewidth}{lllX}
      \toprule
      \rowcolor{white}
      Task & Dataset & Split sizes & Example (label) \\
      \midrule
      \texttt{SENT} & SST-2 &67k/872 &a sweet and modest and ultimately winning story (positive) \\
  & Suite &89k/44k/44k &I thought the aircraft would be beautiful, but it wasn't (negative). \\
  \midrule
  \texttt{PARA} & QQP &363k/40k &Q1: What is best way to reach Kashmir / Srinagar? Q2: What is your review of Srinagar, Jammu \& Kashmir, India? (Not duplicate) \\
  & Suite &103k/51k/51k &Q1: How can I become a powerless person? Q2: How can I become a person who is not powerful? (Duplicate) \\
  \midrule
  \texttt{READ} & SQuAD &87k/10k &C: After Hurricane Katrina in 2005, Beyoncé and Rowland founded the Survivor Foundation to provide transitional housing for victims in the Houston area [...] Q: What foundation did Beyoncé start after Hurricane Katrina? (Survivor Foundation) \\
  & Suite &35k/17k/17k &C: Kevin is nicer than Amanda. Q: Who is less nice? (Amanda) \\
  \midrule
  \texttt{HATE} & Davidson &19k/2k/2k &[USER] can a quote this and tag a bitch (Not hateful) \\
  & Founta &79k/10k/10k &rt [USER]: i'm tired of u feminist bitches bc this is just disgusting [URL] (hateful) \\
  & Suite &1.8k/920/921 &It's disheartening to still see people call for the death of women in 2020. (not hateful) \\
      \bottomrule
     \end{tabularx}
     \caption{Summary of the datasets and suites used in this work. We report train/validation/test sizes for the datasets with public test sets and train/validation sizes otherwise. We use the suite splits from \citet{luzdearaujo2023cross,luzdearaujo2022checking}.}
     \label{tab:dataStatistics}
  \end{table*}
\subsection{Tasks}
\label{sec:tasks}
We examine four NLP tasks based on the availability of test suites, their representativeness of the NLP field, and their use in previous work on functionality learning.
All the data is in English.
Table~\ref{tab:dataStatistics} provides examples and split sizes for all datasets and suites.

\textbf{Sentiment analysis (\texttt{SENT})}: the task is to output the sentiment of the input sentence.
The dataset is the Stanford Sentiment Treebank (SST\nobreakdash2) \cite{socher2013recursive}, as made available in the GLUE benchmark \cite{wang2018glue}.
We use the sentiment analysis suite developed by \citet{ribeiro2020accuracy} as the suite.
There is a label space discrepancy between dataset and suite: the dataset labels include only positive and negative, while the suite extends the options with a neutral label.

\textbf{Paraphrase identification (\texttt{PARA})}: the task is to assess if two questions have the same meaning.
We use Quora Question Pairs (QQP) \cite{iyer2017quora} as the dataset and the QQP suite by \citet{ribeiro2020accuracy}.

\textbf{Reading comprehension (\texttt{READ})}: given a context paragraph, the task is to answer a question whose answer is in the context.
We use the Stanford Question Answering Dataset (SQuAD) \cite{rajpurkar2016squad} as the dataset and the corresponding suite by \citet{ribeiro2020accuracy}.

\textbf{Hate speech detection (\texttt{HATE})}: the task is to determine whether a given sentence contains hateful speech.
Following previous work \cite{rottger2021hatecheck}, we examine two datasets \cite{davidson2017automated,founta2018large}, which we refer to as Davidson and Founta.
We use \hatecheck \cite{rottger2021hatecheck} as the suite.

\subsection{Models}
We compare the predictions of all models from the Flan-T5 family \cite{chung2022scaling,wei2022finetuned} (Small, Base, Large, XL and XXL), Zephyr \cite{tunstall2023Zephyr}, and ChatGPT\footnote{The \texttt{gpt-3.5-turbo-0301} variant of the OpenAI API.} \cite{openai2022chatGPT}.
To examine the model size effect, we cover several orders of magnitude---from 80M to billions of parameters.\footnote{From smallest to largest: Small-80M , Base-250M, Large-780M, XL-3B, Zephyr-7B, and XXl-11B. OpenAI has not disclosed details for GPT-3.5, but the largest variant of its ``sibling model''\cite{openai2022chatGPT} InstructGPT, has 175B parameters.}
These three model families cover three of the main paradigms of LLMs---Flan-T5 are instruction-tuned models~\cite{longpre2023Flan}, Zephyr is a chat model aligned with human preferences through direct preference optimization (DPO) \cite{rafailov2023direct}, while ChatGPT is aligned through reinforcement learning from human feedback (RLHF)~\cite{openai2022chatGPT}.

\subsection{Specification instruction generation}
\label{sec:ruleGen}
We experiment with handcrafted and machine-generated specification instructions.
Tables \ref{tab:sentFuncs}-\ref{tab:hateFuncs} in Appendix~\ref{sec:rules} exhibit all specification instructions from both settings.

\textbf{Handcrafted.}
The specification instructions in the handcrafted setting were manually written by one of the authors.
Specification instructions for the \checklist suites (\texttt{SENT}, \texttt{PARA} and \texttt{READ}) were freely written based on the functionalities in the suite.
This was done by manual inspection of each functionality's test cases and documentation.
Since \hatecheck contains natural language descriptions of all functionalities \cite[Appendix B]{rottger2021hatecheck}, we adapt them to fit our specification format.

\textbf{Machine-generated.}
We designed a prompt template in which we provide the task, the functionality name,\footnote{Names were taken from the suite for the \checklist suites or \citet{rottger2021hatecheck} for \hatecheck.} six\footnote{Two, in the case of \texttt{READ} INV functionalities, due to its lengthy inputs.} test cases from the functionality and ask for a rule that supports the behavior encoded by the test cases. Table~\ref{tab:ruleGenPrompts} in Appendix~\ref{sec:genPrompts} shows an example for each task-test type combination.
We then generated a prompt for each functionality, fed it to ChatGPT, and used the completions as the machine-generated specification instructions.

\subsection{Evaluation metrics}
\label{sec:eval}
\textbf{Dataset metrics}:
We use the accuracy as the metric for SST-2 and QQP, the exact string match for SQuAD, and the \fscore score of the hateful class for Founta and Davidson.\footnote{We use Scikit-learn \cite{scikit-learn} to compute the \fscore score and \hface Datasets \cite{lhoest2021datasets} for the other metrics.}

\textbf{Suite metrics}:
Each functionality has a pass rate: the fraction of successful test cases.
The final suite score is the arithmetic mean of all its functionality pass rates.
Each evaluation scenario (§ \ref{sec:crossFunc}) yields a corresponding suite score.
Therefore, a suite has (1) seen, (2) functionality, and (3) functionality class generalization scores. 

\textbf{Aggregate metrics}:
We report the generalization score \g \cite{luzdearaujo2023cross} as the aggregate score of suite and dataset performance.
It is the harmonic mean of the dataset and suite metrics.
The harmonic mean is used so that high dataset performance cannot compensate for poor suite performance (and vice-versa).
Each suite metric yields its own aggregate score: \gseen, \gfunc, and \gclass for seen, functionality generalization, and functionality class generalization.\footnote{When using the baseline prompting methods (with no specifications), the three \g scores are the same, as specification instructions are never included in the prompt.}

\textbf{Evaluation of machine-generated specification instructions}:
We manually evaluate the quality of the ChatGPT-generated specification instructions using the criteria established by \citet{wang2023SelfInstruct}, where each generated specification instruction is assigned a rating from A (best) to D (worst).\footnote{Details in Appendix~\ref{sec:rules}.}

\begin{figure*}[tb]
  \centering
  \includegraphics[width=\linewidth]{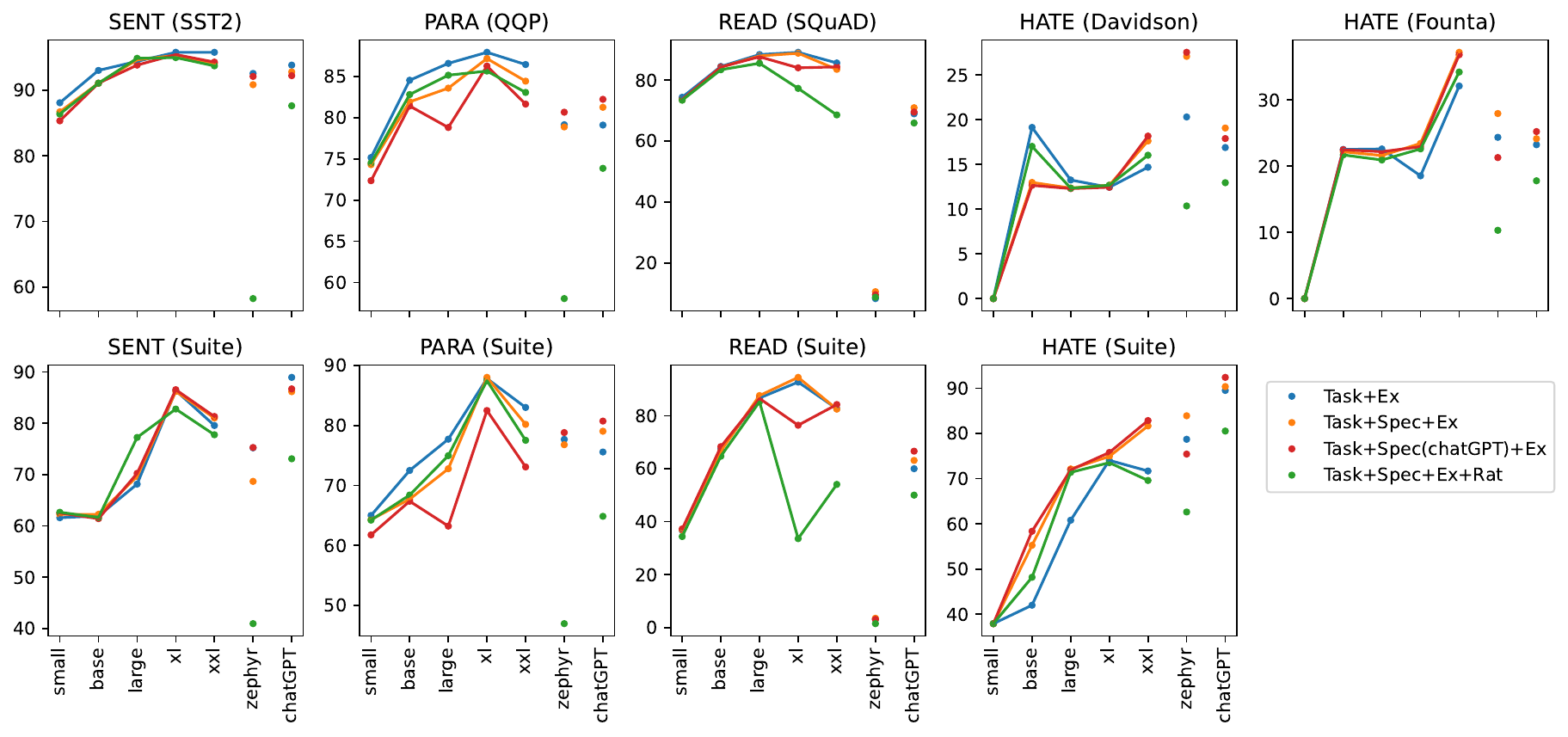}
  \caption{Dataset and suite results for exemplar-augmented prompts. Results for prompts without exemplars are shown in Appendix~\ref{sec:addResults}. Results from the Flan-T5 models are connected with lines to denote that they share the same architecture, training data and training procedure, varying only in number of parameters~\cite{chung2022scaling}.}
  \label{fig:dataAndSuite}
\end{figure*}


\begin{table*}[tb]
    \scriptsize
    \centering
   \renewcommand{\arraystretch}{.8}%
    \begin{tabularx}{\linewidth}{XlXXXXXXXXXXXXXXXX}
      \toprule
      \multirow{2}{*}{\rotatebox[origin=c]{90}{Model}} & Method & \multicolumn{3}{c}{\texttt{SENT}} & \multicolumn{3}{c}{\texttt{PARA}} & \multicolumn{3}{c}{\texttt{READ}} & \multicolumn{3}{c}{\texttt{HATE}-D} & \multicolumn{3}{c}{\texttt{HATE}-F} & \\
      \cmidrule(lr){3-5}\cmidrule(lr){6-8}\cmidrule(lr){9-11}\cmidrule(lr){12-14}\cmidrule(lr){15-17}
      &  &  \gseen & \gfunc & \gclass  & \gseen & \gfunc & \gclass  & \gseen & \gfunc & \gclass  & \gseen & \gfunc & \gclass  & \gseen & \gfunc & \gclass & Avg.\\
      \midrule
      \multirow{6}{*}{\rotatebox[origin=c]{90}{Small}} & Task & 72.65 & 72.65 & 72.65 & 71.12 & 71.12 & 71.12 & 48.96 & 48.96 & 48.96 & 0.00 & 0.00 & 0.00 & 0.00 & 0.00 & 0.00 & 38.55 \\
      & Task+Ex & 72.49 & 72.49 & 72.49 & 69.71 & 69.71 & 69.71 & 49.19 & 49.19 & 49.19 & 0.00 & 0.00 & 0.00 & 0.00 & 0.00 & 0.00 & 38.28 \\
      \cmidrule{2-18}
      & Task+Spec & 72.65 & 72.52 & 72.58 & \textcolor{red}{67.71} & \textcolor{red}{67.22} & \textcolor{red}{67.84} & \textcolor{red}{45.84} & \textcolor{red}{46.42} & \textcolor{red}{46.91} & 0.00 & 0.00 & 0.00 & 0.00 & 0.00 & 0.00 & \textcolor{red}{37.31} \\
      & Task+Spec+Ex & 72.47 & 72.45 & 72.45 & \textcolor{red}{68.94} & \textcolor{red}{68.87} & \textcolor{red}{68.87} & 49.16 & 49.65 & 49.76 & 0.00 & 0.00 & 0.00 & 0.00 & 0.00 & 0.00 & 38.17 \\
      & Task+Spec(chatGPT)+Ex & 72.16 & 72.18 & 72.31 & \textcolor{red}{66.63} & \textcolor{red}{66.53} & \textcolor{red}{67.82} & 49.48 & 49.93 & \textcolor{PineGreen}{50.28} & 0.00 & 0.00 & 0.00 & 0.00 & 0.00 & 0.00 & \textcolor{red}{37.82} \\
      & Task+Spec+Rat & 72.27 & 72.14 & 72.11 & \textcolor{red}{66.97} & \textcolor{red}{66.98} & \textcolor{red}{67.12} & \textcolor{red}{29.43} & \textcolor{red}{29.48} & \textcolor{red}{29.94} & 0.00 & 0.00 & 0.00 & 0.00 & 0.00 & 0.00 & \textcolor{red}{33.76} \\
      & Task+Spec+Ex+Rat & 72.62 & 72.60 & 72.25 & \textcolor{red}{68.98} & \textcolor{red}{68.94} & \textcolor{red}{68.88} & \textcolor{red}{46.84} & \textcolor{red}{47.33} & \textcolor{red}{47.41} & 0.00 & 0.00 & 0.00 & 0.00 & 0.00 & 0.00 & \textcolor{red}{37.72} \\
   \midrule
  \multirow{5}{*}{\rotatebox[origin=c]{90}{Base}} & Task & 74.75 & 74.75 & 74.75 & 77.50 & 77.50 & 77.50 & 75.37 & 75.37 & 75.37 & 22.03 & 22.03 & 22.03 & 29.89 & 29.89 & 29.89 & 55.91 \\
  & Task+Ex & 74.36 & 74.36 & 74.36 & 78.05 & 78.05 & 78.05 & 74.98 & 74.98 & 74.98 & 26.30 & 26.30 & 26.30 & 29.32 & 29.32 & 29.32 & 56.60 \\
  \cmidrule{2-18}
  & Task+Spec & \textcolor{PineGreen}{76.01} & \textcolor{PineGreen}{76.38} & \textcolor{PineGreen}{75.38} & \textcolor{red}{72.96} & \textcolor{red}{72.98} & \textcolor{red}{72.99} & \textcolor{red}{72.75} & \textcolor{red}{72.91} & \textcolor{red}{72.93} & 19.98 & \textcolor{red}{19.94} & \textcolor{red}{19.68} & 29.02 & 28.95 & 28.39 & \textcolor{red}{54.08} \\
  & Task+Spec+Ex & 73.95 & 73.99 & \textcolor{red}{73.57} & \textcolor{red}{74.14} & \textcolor{red}{74.14} & \textcolor{red}{74.15} & \textcolor{red}{74.44} & 74.59 & \textcolor{red}{74.26} & \textcolor{red}{21.03} & \textcolor{red}{21.03} & \textcolor{red}{20.97} & \textcolor{PineGreen}{31.64} & \textcolor{PineGreen}{31.62} & \textcolor{PineGreen}{31.49} & \textcolor{red}{55.00} \\
  & Task+Spec(chatGPT)+Ex & \textcolor{red}{73.35} & 73.96 & \textcolor{red}{73.14} & \textcolor{red}{73.73} & \textcolor{red}{73.54} & \textcolor{red}{73.55} & 75.47 & \textcolor{PineGreen}{75.72} & 75.04 & \textcolor{red}{20.80} & \textcolor{red}{20.80} & \textcolor{red}{20.75} & \textcolor{PineGreen}{32.44} & \textcolor{PineGreen}{32.42} & \textcolor{PineGreen}{32.30} & \textcolor{red}{55.13} \\
  & Task+Spec+Rat & \textcolor{PineGreen}{77.10} & \textcolor{PineGreen}{77.05} & \textcolor{PineGreen}{76.52} & \textcolor{red}{74.14} & \textcolor{red}{74.27} & \textcolor{red}{74.13} & \textcolor{red}{65.22} & \textcolor{red}{67.31} & \textcolor{red}{67.80} & \textcolor{PineGreen}{24.36} & \textcolor{PineGreen}{24.32} & 24.07 & \textcolor{red}{26.63} & \textcolor{red}{26.58} & \textcolor{red}{26.29} & \textcolor{red}{53.72} \\
  & Task+Spec+Ex+Rat & \textcolor{red}{73.52} & 73.76 & \textcolor{red}{73.19} & \textcolor{red}{74.92} & \textcolor{red}{74.90} & \textcolor{red}{74.91} & \textcolor{red}{72.88} & \textcolor{red}{72.88} & \textcolor{red}{72.78} & 25.16 & 25.12 & 25.00 & 29.91 & 29.85 & 29.69 & \textcolor{red}{55.23} \\
   \midrule
  \multirow{5}{*}{\rotatebox[origin=c]{90}{Large}} & Task & 80.21 & 80.21 & 80.21 & 81.79 & 81.79 & 81.79 & 87.03 & 87.03 & 87.03 & 20.67 & 20.67 & 20.67 & 30.07 & 30.07 & 30.07 & 59.95 \\
  & Task+Ex & 79.15 & 79.15 & 79.15 & 81.91 & 81.91 & 81.91 & 87.53 & 87.53 & 87.53 & 21.77 & 21.77 & 21.77 & 32.94 & 32.94 & 32.94 & 60.66 \\
  \cmidrule{2-18}
  & Task+Spec & \textcolor{PineGreen}{83.99} & \textcolor{PineGreen}{84.06} & \textcolor{PineGreen}{83.41} & \textcolor{red}{72.06} & \textcolor{red}{71.87} & \textcolor{red}{71.76} & 87.34 & \textcolor{PineGreen}{87.64} & \textcolor{PineGreen}{87.67} & 21.11 & 21.11 & 21.11 & \textcolor{PineGreen}{32.49} & \textcolor{PineGreen}{32.51} & \textcolor{PineGreen}{32.49} & \textcolor{red}{59.37} \\
  & Task+Spec+Ex & \textcolor{PineGreen}{80.23} & \textcolor{PineGreen}{80.08} & \textcolor{PineGreen}{79.70} & \textcolor{red}{77.81} & \textcolor{red}{77.82} & \textcolor{red}{77.45} & 87.76 & 87.81 & 87.77 & \textcolor{red}{21.11} & \textcolor{red}{21.10} & \textcolor{red}{21.10} & 33.24 & 33.22 & 33.23 & \textcolor{red}{59.96} \\
  & Task+Spec(chatGPT)+Ex & \textcolor{PineGreen}{80.33} & \textcolor{PineGreen}{80.46} & 79.50 & \textcolor{red}{70.15} & \textcolor{red}{69.57} & \textcolor{red}{69.54} & 87.08 & 87.24 & 87.14 & \textcolor{red}{21.01} & \textcolor{red}{21.00} & \textcolor{red}{20.99} & 33.90 & 33.88 & 33.85 & \textcolor{red}{58.38} \\
  & Task+Spec+Rat & \textcolor{PineGreen}{82.54} & \textcolor{PineGreen}{82.29} & \textcolor{PineGreen}{81.24} & \textcolor{red}{68.08} & \textcolor{red}{68.03} & \textcolor{red}{68.31} & \textcolor{red}{28.87} & \textcolor{red}{28.81} & \textcolor{red}{27.97} & \textcolor{PineGreen}{21.28} & \textcolor{PineGreen}{21.28} & \textcolor{PineGreen}{21.25} & \textcolor{PineGreen}{34.14} & \textcolor{PineGreen}{34.14} & \textcolor{PineGreen}{34.06} & \textcolor{red}{46.82} \\
  & Task+Spec+Ex+Rat & \textcolor{PineGreen}{85.15} & \textcolor{PineGreen}{85.33} & \textcolor{PineGreen}{85.45} & \textcolor{red}{79.73} & \textcolor{red}{79.68} & \textcolor{red}{79.49} & \textcolor{red}{85.41} & \textcolor{red}{85.42} & \textcolor{red}{85.43} & 21.05 & 21.09 & 21.06 & 32.36 & 32.45 & 32.38 & 60.77 \\
   \midrule
  \multirow{5}{*}{\rotatebox[origin=c]{90}{XL}} & Task & 90.76 & 90.76 & 90.76 & \textbf{87.89} & \textbf{87.89} & \textbf{87.89} & 89.04 & 89.04 & 89.04 & 20.85 & 20.85 & 20.85 & 29.44 & 29.44 & 29.44 & 63.60 \\
  & Task+Ex & 90.83 & 90.83 & 90.83 & 87.88 & 87.88 & 87.88 & 90.88 & 90.88 & 90.88 & 21.31 & 21.31 & 21.31 & 29.65 & 29.65 & 29.65 & 64.11 \\
  \cmidrule{2-18}
  & Task+Spec & 89.85 & 89.91 & \textcolor{red}{89.53} & \textcolor{red}{84.34} & \textcolor{red}{84.32} & \textcolor{red}{84.41} & \textcolor{PineGreen}{90.99} & \textcolor{PineGreen}{\textbf{91.27}} & \textcolor{PineGreen}{90.84} & \textcolor{PineGreen}{21.63} & \textcolor{PineGreen}{21.61} & \textcolor{PineGreen}{21.64} & \textcolor{PineGreen}{32.16} & \textcolor{PineGreen}{32.12} & \textcolor{PineGreen}{32.19} & 63.79 \\
  & Task+Spec+Ex & 90.53 & 90.44 & \textcolor{red}{90.10} & \textcolor{red}{87.59} & \textcolor{red}{87.46} & \textcolor{red}{87.09} & \textcolor{PineGreen}{\textbf{91.56}} & 91.23 & \textbf{91.05} & 21.69 & 21.69 & 21.67 & \textcolor{PineGreen}{35.69} & \textcolor{PineGreen}{35.70} & \textcolor{PineGreen}{35.66} & \textcolor{PineGreen}{65.28} \\
  & Task+Spec(chatGPT)+Ex & 90.77 & 90.37 & 90.62 & \textcolor{red}{84.34} & \textcolor{red}{84.28} & \textcolor{red}{84.31} & \textcolor{red}{80.04} & \textcolor{red}{80.63} & \textcolor{red}{79.62} & 21.38 & 21.37 & 21.40 & \textcolor{PineGreen}{35.21} & \textcolor{PineGreen}{35.18} & \textcolor{PineGreen}{35.24} & \textcolor{red}{62.32} \\
  & Task+Spec+Rat & \textcolor{red}{86.61} & \textcolor{red}{86.39} & \textcolor{red}{87.76} & \textcolor{red}{84.10} & \textcolor{red}{84.22} & \textcolor{red}{84.43} & \textcolor{red}{27.94} & \textcolor{red}{29.67} & \textcolor{red}{30.90} & 21.16 & 21.14 & 21.17 & \textcolor{PineGreen}{31.25} & \textcolor{PineGreen}{31.21} & \textcolor{PineGreen}{31.28} & \textcolor{red}{50.62} \\
  & Task+Spec+Ex+Rat & \textcolor{red}{88.45} & \textcolor{red}{88.09} & \textcolor{red}{88.67} & \textcolor{red}{86.55} & \textcolor{red}{86.41} & \textcolor{red}{86.54} & \textcolor{red}{46.83} & \textcolor{red}{47.19} & \textcolor{red}{48.00} & 21.63 & 21.60 & 21.65 & \textcolor{PineGreen}{34.57} & \textcolor{PineGreen}{34.51} & \textcolor{PineGreen}{34.62} & \textcolor{red}{55.69} \\
   \midrule
  \multirow{4}{*}{\rotatebox[origin=c]{90}{XXL}} &Task & 82.13 & 82.13 & 82.13 & 84.51 & 84.51 & 84.51 & 82.07 & 82.07 & 82.07 & 22.26 & 22.26 & 22.26 & 36.82 & 36.82 & 36.82 & 61.56 \\
  & Task+Ex & 86.92 & 86.92 & 86.92 & 84.69 & 84.69 & 84.69 & 84.12 & 84.12 & 84.12 & 24.39 & 24.39 & 24.39 & 44.33 & 44.33 & 44.33 & 64.89 \\
  \cmidrule{2-18}
  & Task+Spec & \textcolor{PineGreen}{84.08} & 82.94 & 82.84 & \textcolor{red}{76.43} & \textcolor{red}{76.08} & \textcolor{red}{76.52} & \textcolor{red}{81.06} & 81.57 & \textcolor{red}{80.77} & \textcolor{PineGreen}{27.85} & \textcolor{PineGreen}{27.77} & \textcolor{PineGreen}{27.80} & \textcolor{PineGreen}{47.88} & \textcolor{PineGreen}{47.64} & \textcolor{PineGreen}{47.74} & \textcolor{PineGreen}{63.27} \\
  & Task+Spec+Ex & 87.05 & \textcolor{red}{85.73} & \textcolor{red}{85.74} & \textcolor{red}{82.26} & \textcolor{red}{82.23} & \textcolor{red}{82.28} & \textcolor{red}{83.01} & \textcolor{red}{82.21} & \textcolor{red}{82.67} & \textcolor{PineGreen}{29.00} & \textcolor{PineGreen}{29.07} & \textcolor{PineGreen}{28.93} & \textcolor{PineGreen}{\textbf{51.09}} & \textcolor{PineGreen}{\textbf{51.31}} & \textcolor{PineGreen}{\textbf{50.87}} & \textcolor{PineGreen}{\textbf{66.23}} \\
  & Task+Spec(chatGPT)+Ex & 87.32 & 86.40 & \textcolor{PineGreen}{88.31} & \textcolor{red}{77.13} & \textcolor{red}{76.62} & \textcolor{red}{76.65} & 84.28 & 83.71 & \textcolor{red}{83.32} & \textcolor{PineGreen}{29.80} & \textcolor{PineGreen}{29.73} & \textcolor{PineGreen}{29.77} & \textcolor{PineGreen}{50.97} & \textcolor{PineGreen}{50.74} & \textcolor{PineGreen}{50.86} & \textcolor{PineGreen}{65.71} \\
  & Task+Spec+Rat & 83.90 & 83.16 & 82.51 & \textcolor{red}{72.44} & \textcolor{red}{72.70} & \textcolor{red}{72.05} & \textcolor{red}{11.30} & \textcolor{red}{11.51} & \textcolor{red}{11.29} & \textcolor{PineGreen}{30.33} & \textcolor{PineGreen}{30.25} & \textcolor{PineGreen}{29.99} & \textcolor{PineGreen}{48.02} & \textcolor{PineGreen}{47.84} & \textcolor{PineGreen}{47.19} & \textcolor{red}{48.97} \\
  & Task+Spec+Ex+Rat & \textcolor{red}{84.99} & \textcolor{red}{83.11} & \textcolor{red}{84.59} & \textcolor{red}{80.20} & \textcolor{red}{79.96} & \textcolor{red}{80.18} & \textcolor{red}{60.47} & \textcolor{red}{60.50} & \textcolor{red}{60.40} & \textcolor{PineGreen}{26.06} & 25.83 & 25.82 & 45.86 & 45.16 & 45.10 & \textcolor{red}{59.21} \\
   \midrule
   \multirow{4}{*}{\rotatebox[origin=c]{90}{Zephyr}} &Task & 89.18 & 89.18 & 89.18 & 63.32 & 63.32 & 63.32 & 1.32 & 1.32 & 1.32 & 28.73 & 28.73 & 28.73 & 33.64 & 33.64 & 33.64 & 43.24 \\
   & Task+Ex & 82.99 & 82.99 & 82.99 & 78.41 & 78.41 & 78.41 & 4.60 & 4.60 & 4.60 & 32.29 & 32.29 & 32.29 & 37.19 & 37.19 & 37.19 & 47.09 \\
   \cmidrule{2-18}
   & Task+Spec & \textcolor{red}{87.13} & \textcolor{red}{85.67} & \textcolor{red}{86.30} & \textcolor{red}{61.31} & \textcolor{red}{60.76} & \textcolor{red}{61.97} & \textcolor{PineGreen}{8.55} & \textcolor{PineGreen}{8.68} & \textcolor{PineGreen}{8.63} & \textcolor{PineGreen}{31.31} & \textcolor{PineGreen}{31.19} & \textcolor{PineGreen}{31.20} & \textcolor{PineGreen}{44.11} & \textcolor{PineGreen}{43.88} & \textcolor{PineGreen}{43.88} & \textcolor{PineGreen}{46.30} \\
   & Task+Spec+Ex & \textcolor{red}{78.22} & \textcolor{red}{77.25} & \textcolor{red}{77.83} & \textcolor{red}{77.82} & \textcolor{red}{77.83} & \textcolor{red}{77.66} & 5.29 & 5.22 & 5.15 & \textcolor{PineGreen}{\textbf{40.95}} & \textcolor{PineGreen}{\textbf{40.85}} & \textcolor{PineGreen}{\textbf{40.95}} & \textcolor{PineGreen}{41.91} & \textcolor{PineGreen}{41.81} & \textcolor{PineGreen}{41.91} & \textcolor{PineGreen}{48.71} \\
   & Task+Spec(chatGPT)+Ex & 82.84 & \textcolor{red}{80.92} & \textcolor{red}{79.12} & \textcolor{PineGreen}{79.73} & \textcolor{PineGreen}{79.02} & \textcolor{PineGreen}{80.04} & 4.66 & 4.84 & 4.84 & \textcolor{PineGreen}{40.35} & \textcolor{PineGreen}{40.29} & \textcolor{PineGreen}{40.38} & \textcolor{red}{33.20} & \textcolor{red}{33.17} & \textcolor{red}{33.23} & 47.78 \\
   & Task+Spec+Rat & \textcolor{red}{70.39} & \textcolor{red}{69.52} & \textcolor{red}{70.64} & \textcolor{red}{61.26} & \textcolor{red}{59.67} & \textcolor{red}{57.74} & \textcolor{PineGreen}{3.16} & \textcolor{PineGreen}{3.19} & \textcolor{PineGreen}{3.13} & 27.63 & 27.59 & 27.57 & \textcolor{PineGreen}{42.14} & \textcolor{PineGreen}{42.06} & \textcolor{PineGreen}{42.01} & \textcolor{red}{40.51} \\
   & Task+Spec+Ex+Rat & \textcolor{red}{48.10} & \textcolor{red}{47.91} & \textcolor{red}{48.66} & \textcolor{red}{51.91} & \textcolor{red}{51.98} & \textcolor{red}{52.97} & \textcolor{red}{2.53} & \textcolor{red}{2.00} & \textcolor{red}{1.79} & \textcolor{red}{17.78} & \textcolor{red}{17.85} & \textcolor{red}{17.82} & \textcolor{red}{17.70} & \textcolor{red}{17.77} & \textcolor{red}{17.74} & \textcolor{red}{27.63} \\
   \midrule
  \multirow{5}{*}{\rotatebox[origin=c]{90}{ChatGPT}} & Task & \textbf{93.07} & \textbf{93.07} & \textbf{93.07} & 74.81 & 74.81 & 74.81 & 14.39 & 14.39 & 14.39 & 23.53 & 23.53 & 23.53 & 38.63 & 38.63 & 38.63 & 48.89 \\
  & Task+Ex & 91.32 & 91.32 & 91.32 & 77.30 & 77.30 & 77.30 & 64.17 & 64.17 & 64.17 & 28.42 & 28.42 & 28.42 & 36.88 & 36.88 & 36.88 & 59.62 \\
  \cmidrule{2-18}
  & Task+Spec & \textcolor{red}{89.40} & \textcolor{red}{87.05} & \textcolor{red}{88.77} & \textcolor{PineGreen}{75.50} & \textcolor{red}{74.03} & \textcolor{red}{73.69} & \textcolor{PineGreen}{19.30} & \textcolor{PineGreen}{19.71} & \textcolor{PineGreen}{20.79} & \textcolor{PineGreen}{24.72} & \textcolor{PineGreen}{24.67} & \textcolor{PineGreen}{24.67} & \textcolor{PineGreen}{43.23} & \textcolor{PineGreen}{43.07} & \textcolor{PineGreen}{43.05} & \textcolor{PineGreen}{50.11} \\
  & Task+Spec+Ex & \textcolor{red}{89.35} & \textcolor{red}{87.94} & \textcolor{red}{90.27} & \textcolor{PineGreen}{80.13} & \textcolor{PineGreen}{78.01} & \textcolor{PineGreen}{78.78} & \textcolor{PineGreen}{66.81} & \textcolor{PineGreen}{67.04} & \textcolor{PineGreen}{66.74} & \textcolor{PineGreen}{31.50} & \textcolor{PineGreen}{31.46} & \textcolor{PineGreen}{31.46} & 38.09 & 38.03 & 38.03 & \textcolor{PineGreen}{60.91} \\
  & Task+Spec(chatGPT)+Ex & \textcolor{red}{89.37} & \textcolor{red}{88.19} & \textcolor{red}{89.65} & \textcolor{PineGreen}{81.47} & \textcolor{PineGreen}{79.53} & \textcolor{PineGreen}{82.11} & \textcolor{PineGreen}{68.04} & \textcolor{PineGreen}{66.06} & \textcolor{PineGreen}{65.47} & \textcolor{PineGreen}{29.99} & \textcolor{PineGreen}{29.94} & \textcolor{PineGreen}{29.87} & \textcolor{PineGreen}{39.62} & \textcolor{PineGreen}{39.54} & \textcolor{PineGreen}{39.42} & \textcolor{PineGreen}{61.22} \\
  & Task+Spec+Rat & \textcolor{red}{78.04} & \textcolor{red}{75.27} & \textcolor{red}{75.25} & \textcolor{red}{65.83} & \textcolor{red}{64.03} & \textcolor{red}{64.60} & \textcolor{red}{8.50} & \textcolor{red}{8.05} & \textcolor{red}{7.67} & \textcolor{red}{21.41} & \textcolor{red}{21.28} & \textcolor{red}{21.33} & \textcolor{red}{33.59} & \textcolor{red}{33.29} & \textcolor{red}{33.39} & \textcolor{red}{40.77} \\
  & Task+Spec+Ex+Rat & \textcolor{red}{79.69} & \textcolor{red}{78.52} & \textcolor{red}{80.75} & \textcolor{red}{69.06} & \textcolor{red}{66.96} & \textcolor{red}{66.35} & \textcolor{red}{56.89} & \textcolor{red}{55.42} & \textcolor{red}{56.24} & \textcolor{red}{22.33} & \textcolor{red}{22.12} & \textcolor{red}{22.16} & \textcolor{red}{29.13} & \textcolor{red}{28.79} & \textcolor{red}{28.85} & \textcolor{red}{50.88} \\
      \bottomrule
     \end{tabularx}
     \caption{Suite and dataset aggregate scores (in \%).
     \texttt{HATE-D} and \texttt{HATE-F} indicate the aggregate scores for using Davidson and Founta as the dataset.
     Scores significantly above or below the corresponding baseline (Task and Task+Ex for prompts without and with examplars) are shown in green and red respectively. The best score for each measure is highlighted in bold weight. Scores not significantly different from the baseline are shown in black.}
     \label{tab:aggResults}
  \end{table*}

\begin{figure*}[tb]
  \centering
  \includegraphics[width=\linewidth]{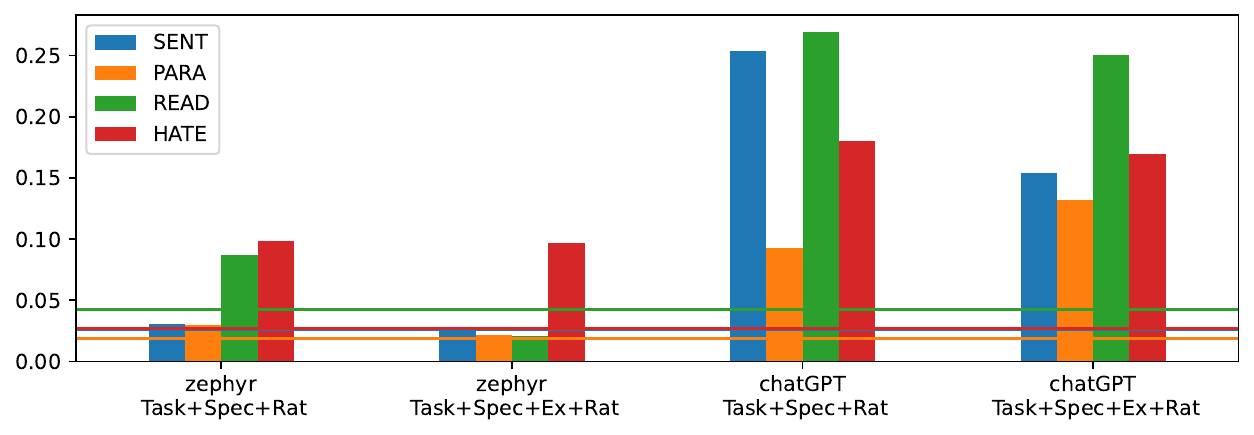}
  \caption{Specification prediction \fscore scores. The horizontal lines show results for a classifier that randomly selects a specification.}
  \label{fig:ruleEval}
\end{figure*}

\begin{table}[tb]
  \footnotesize
  \centering
  \begin{tabularx}{\linewidth}{Xrrrr}
    \toprule
    & \multicolumn{2}{X}{Func.-wise} & \multicolumn{2}{X}{Instance-wise} \\
    \cmidrule(lr){2-3}\cmidrule(lr){4-5}
    Task &-Ex &+Ex &-Ex &+Ex \\
    \midrule
    \texttt{SENT} &0.36 &0.30 &0.26 &0.14 \\
    \texttt{PARA} &0.27 &0.23 &0.25 &0.16 \\
    \texttt{READ} &-0.39 &0.19 &-0.20 &0.09 \\
    \texttt{HATE} &0.23 &0.19 &0.19 &0.12 \\
    \bottomrule
   \end{tabularx}
   \caption{Pearson's correlations between specification prediction and task performance on the functionality-aggregate level and instance-wise.
   -Ex and +Ex indicate if the prompt includes a Exemplars module.}
   \label{tab:corrs}
\end{table}

\section{Analysis of prompt methods and components}
Table~\ref{tab:aggResults} shows the aggregate scores for all methods and models.\footnote{We report scores for a single run. We test for significance through randomized testing \cite{yeh2000accurate} (10000 rounds, $p < 0.05$). We report all p-values in Table~\ref{tab:pvalues}.}
Fig.~\ref{fig:dataAndSuite} shows scores for all suites and datasets.\footnote{Appendix~\ref{sec:length} analyses the impact of prompt length on performance and Appendix~\ref{sec:addResults} shows scores for each suite and dataset.}





\textbf{Impact of specification instructions.}
Specification instructions only improved performance of the larger models: including them in the prompt reduced the average performance of Flan-T5-Large and smaller models but improved it for Flan-T5-XL and larger models.
The effect differed across tasks and no model-method pair improved over the baseline in all tasks.
ChatGPT benefitted from specification instructions most consistently (four out of five tasks).

We expected specification instructions to improve suites' scores (Fig.~\ref{fig:dataAndSuite}, bottom row) more than datasets (top row), because specification instructions are guaranteed to correspond to suite instances and only occasionally to dataset instances.
We validate this intuition by comparing the average dataset and average suite performance difference between specification-augmented prompts and their corresponding baseline (Fig.~\ref{fig:dataDiffsAgg} in Appendix~\ref{sec:addResults}).
While Flan-T5-base and larger models benefit from specification instructions considering suite performance, only XL and larger models could improve dataset performance. 
Table~\ref{tab:qualEval} in Appendix~\ref{sec:addResults} shows, for each dataset, the examples for which specification instructions consistently improved or harmed ChatGPT predictions.

\textbf{Impact of exemplars.}
Adding exemplars improved average performance in almost all the scenarios.
The only exception was for Flan-T5-small baseline methods, where Task outperformed Task+Ex by $0.27$.
This effect was overall consistent across tasks: Task+Ex achieved an (averaged across models) improvement over Task in all evaluation scenarios, except for \texttt{SENT}, for which there was an average decrease of $0.67$.
Comparing Task+Spec and Task+Spec+Rat with their exemplar-augmented counterparts yields similar conclusions, except that Task+Spec+Rat outperformed Task+Spec+Ex+Rat on the \texttt{HATE} tasks.

\textbf{Impact of rationales.}
In most cases, average performance decreases when prompts include the Rationale module.
The only exceptions are Flan-T5-Base and Large with Task+Spec+Ex+Rat prompts, outperforming Task+Spec+Ex by 0.23/0.80.
Qualitatively, only Zephyr and ChatGPT actually produced rationales.
Flan-T5-XXL either ignores the rationale instruction or copies the list of specification instructions.
The other models ignore the module entirely.

Even though the Rationale module did not improve task performance, ChatGPT returned the correct specification better than random in all tasks (Fig.~\ref{fig:ruleEval}).
A follow-up question is how much of the performance degradation is due to ChatGPT failing to identify the correct specification.

To investigate this, we computed the Pearson's correlation between specification prediction correctness and functionality performance on two levels: (1) the functionality-aggregate level, where we measure the correlation between functionality pass rates and the corresponding average specification prediction \fscore scores; and (2) the test case level, where we measure the correlation between the specification prediction \fscore score for each test case and a binary score indicating whether ChatGPT passed the test.

Table~\ref{tab:corrs} presents the obtained correlations.The associations between specification prediction and functionality performance were weak (absolute values smaller than $0.4$).
This suggests that the negative impact of the Rationale module can only partially be attributed to it causing ChatGPT to attend to the wrong specifications.
We investigate other reasons for the performance degradation in Appendix~\ref{sec:rat}.

\textbf{ChatGPT vs. human-generated specification instructions.}
Human-written specification instructions led to better average scores than ChatGPT-generated ones: in the majority of the models, Task+Spec(chatGPT)+Ex had a lower average score than Task+Spec+Ex.
ChatGPT itself was one exception, with an absolute improvement of $0.31$ p.p.
That said, using ChatGPT-generated specification instructions still outperformed not using any specifications for the three largest models.
We discuss the quality of the chatGPT-generated specifications in Appendix~\ref{sec:chatGPTrules}.

\textbf{Generalization to unseen functionalities.}
\gseen was frequently strictly higher than \gfunc and \gclass,
indicating a generalization gap between seen and unseen functionalities.
However the score gaps were much less expressive when compared to previous work on functionality learning~\cite{luzdearaujo2023cross,luzdearaujo2022checking,rozen2019diversify}.
These results show that generalization to unseen functionalities---alternatively, overfitting to seen functionalities---is less of a concern here than in previous work.

\section{Analysis of specification interaction}
Specification-augmented prompts include dozens of instructions that can interact with each other to affect model prediction in surprising ways.
To analyze the interaction between specifications, we compare the functionality's pass rates (averaged across models and prompting methods) across the different evaluation scenarios (§~\ref{sec:crossFunc}) and examine the functionalities with the largest improvement and degradation.
That is, each functionality has four pass rates (§~\ref{sec:eval}) for a given model:

\textbf{\sbase}: pass rate when prompts do not include specification instructions.

\textbf{\sseen}: pass rate when prompts include all specification instructions.

\textbf{\sfunc}: pass rate when prompts include all specification instructions minus the one corresponding to the functionality.

\textbf{\sclass}: pass rate when prompts include all specification instructions minus those corresponding to functionalities from the same functionality class.

Pairwise comparison of these scores leads to different insights on specification interactions.
For each possible pair, we rank all functionalities by the score difference (e.g., $\sseen-\sbase$) and examine the functionalities at the ranking extremes.
To support our analysis, we also inspect ChatGPT and Zephyr prediction rationales for examples from the selected functionalities.
We show the model rationales and examples for the extreme functionalities in Table~\ref{tab:funcDiffExs} (App.~\ref{sec:funcDiffsApp}).

\textbf{$\sseen-\sbase$}:
This difference measures how the full set of specification instructions contributes to each functionality score.
Positive and negative differences indicate that the functionality benefitted from or was harmed by the instruction set.
The functionality on the positive extreme is from the \texttt{PARA} suite and states that two identical questions are duplicates even if different irrelevant preambles precede them.
The functionality on the negative extreme was from the same suite and tested for simple pronoun co-reference capabilities.
The ChatGPT rationale show how it applies specification instructions that do not apply to the case and lead to incorrect predictions.
Generally, the most functionalities on the negative extreme require linguistic capabilities (e.g., negation), while the functionalities on the positive extreme described some facet of the task that does not require complex linguistic knowledge (e.g., introducing neutral sentiment in \texttt{SENT}).\footnote{The bottom five functionalities measured negation, antonym and co-reference capabilities, while the top five described neutral sentiment, order invariance of comparisons and how preambles to questions may be irrelevant.}

Rankings obtained from $\sfunc-\sbase$ and $\sclass-\sbase$ yielded the same extreme functionalities and were highly correlated to $\sseen-\sbase$ (Kendall $\tau$ of 0.89 and  0.85 respectively).
That is, functionality pass rates are similar even if one excludes specification instructions corresponding to the functionality (or its class).
The set of specification instructions as a whole plays a bigger role than even the most relevant specification.

\textbf{$\sfunc-\sclass$}:
This measure relates to the interplay between specifications from the same functionality class.
Positive and negative differences indicate constructive and destructive interference between specifications from the same functionality class.
The functionality on the positive extreme is from the \texttt{SENT} suite, and states how sentences using neutral-sentiment words should be neutral.
The rationales illustrate how the model uses specifications from the same class to generate the correct label, unlike the same model with no access to such specifications.
The functionality on the negative extreme posits that a sentence containing a neutral sentiment question with a ``yes'' reply is still neutral. 
The example rationale shows how models mistakenly apply a specification from the same class, which states that replying ``yes'' to a sentiment-laden question affirms the question sentiment.

We discuss the remaining pairs in Appendix \ref{sec:funcDiffsApp}.

\section{Related work}
\textbf{Instruction-following models} This work uses LLMs fine-tuned on instruction data, where tasks are described by natural language instructions \cite{longpre2023Flan,zhou2023LIMA,mishra2022cross,wang2022super}.
Such LLMs have been show to generalize to unseen tasks \cite{muennighoff2023Crosslingual,ouyang2022training,chung2022scaling,wei2022finetuned}.
Our specification instructions differ from traditional instructions: these describe the task (e.g., ``Output the sentiment of the following sentence''). 
In contrast, the specification instructions prescribe the expected behavior for specific cases (e.g., ``the speaker's sentiment should outweigh other opinions'').

\textbf{Instruction induction} Instead of using models to follow instructions, an emerging line of work prompts models to generate instructions \cite{wang2023SelfInstruct,honovich2023Unnatural,honovich2022instruction}.
Our ChatGPT-generated specification instructions can be seen as a form of instruction induction.
An important difference is that previous works prompt the model with input-label pairs and ask it to infer the underlying task.
Our prompts, instead, include the task name and ask the model to infer the underlying labeling rule for the presented exemplars.

\textbf{Model alignment}
An emerging research direction explores how to align LLMs to human values like helpfulness, honesty, and harmlessness \cite{bai2022Constitutional}.
Several approaches have been explored, including fine-tuning models on data constructed to reflect such values \cite{zhou2023LIMA,solaiman2021Process}, and optimizing reward functions derived from human \cite{rafailov2023direct,ganguli2023Capacity,ouyang2022training} or machine-generated \cite{lee2023RLAIF} preferences.
Some works encode human values as a list of rules or principles \cite{sun2023principle,bai2022Constitutional}: natural language sentences that describe the desired values.
Specification instructions align the model not to high-level ethical values but to how a particular task should be performed.

\textbf{Functionality learning}
Previous methods for functionality learning (also called model patching or debugging, behavioral learning, and inoculation) were based on fine-tuning models on functionality data \cite{luzdearaujo2023cross,luzdearaujo2022checking,malon2022fast,murty2022fixing,ribeiro2022adaptive,rozen2019diversify,liu2019inoculation,mccoy2019right}.
That requires constructing new (or holding out) instances for training and additional optimization steps, which can be expensive and unfeasible for large or private models.
Our specification instruction experiments required at most six instances per functionality for machine-generated specification instructions.

\section{Conclusion}
We have studied specification-augmented prompts as a fine-tuning-free way of eliciting LLMs to adopt fine-grained task-specific behaviors.
Our results have shown that specification instructions can improve suite and dataset performance of large models.
That was true for human and ChatGPT-generated specification instructions, though the former were mostly better.
Our cross-functional analysis indicated that improvements are not restricted to the covered functionalities but extend to held-out ones.

Our analysis of specification interaction shows how the specification-augmented prompts' effect differs across functionalities: instructions can help to align models to desired task behaviors (e.g., predicting neutral sentiment) but may deteriorate performance when describing linguistic phenomena.
We show how specifications impact each other in constructive and destructive ways and how the instruction set often leads to the same prediction, even if some specifications relevant to the input are excluded.

Specification-augmented prompts include dozens of instructions, so the predictions result from the interplay of the set of specifications, how they are expressed, the exemplars shown, and the prompt format, among other factors.
Due to the complexity of these matters, rule and principle-based alignment approaches would benefit from interdisciplinary research on how to design and specify rule systems.


\section{Ethical considerations}

Evaluating models on test suites is a valuable technique for finding failure cases and gaining a more comprehensive view of models' capabilities.
However, good scores in a suite may not translate to good performance in the wild, as models may be sensitive to shifts in the data distribution.
Furthermore, suites do not test all relevant aspects of model behavior but merely point out problematic areas only for the specific cases they assess.

We have shown that specification instructions can improve the performance of LLMs, but they are far from being a certificate or guarantee that the model will behave according to them.
Further, while our experiments indicate good generalization, it is still possible that performance on phenomena not covered by the suites has deteriorated (e.g., robustness to adversarial attacks).

\section{Limitations}


Our experiments on specifying functionalities are limited, as we only examine one human-generated set and one machine-generated set of specification instructions.
Specifying a functionality involves many choices, including how to word the instruction, the prompt format, and which specifications should be included.
Each of these is an important aspect that deserves a targeted analysis.

Our results have shown that benefits of specification instructions are task-dependent.
In our experiments, the largest models benefitted from the specification-augmented prompts most consistently, but this may not generalize to other suite-dataset combinations.
Moreover, the datasets and suites we examine are all in English.
A cross-lingual evaluation of specification impact has its own challenges, such as the lack of test suites in lower-resource languages and the matter of how to design specification sets that address the particularities of different languages.

\section*{Acknowledgements}
This research has been funded by the Vienna Science and
Technology Fund (WWTF) [10.47379/VRG19008]
``Knowledge-infused Deep Learning for Natural
Language Processing''.
We are thankful for the credits from the OpenAI API Research Access Program.
We acknowledge EuroHPC Joint Undertaking for awarding us access to MeluXina at LuxProvide, Luxembourg.

\bibliography{refs}

\appendix

\section{Datasets}
\label{sec:datasets}

\paragraph{SST-2} \cite{socher2013recursive} 

\textbf{Data:} movie reviews excerpts from \url{rottentomatoes.com}.

\textbf{Annotation:} Amazon Mechanical Turk workers labeled excerpts with their sentiments.
This work uses the version made available in the GLUE benchmark \cite{wang2018glue}, which provides binary labels for positive/negative sentiment.

\textbf{License:} CC-BY 4.0.

\paragraph{QQP} \cite{iyer2017quora} 

\textbf{Data:} questions pairs from \url{quora.com}.

\textbf{Annotation:} ground truth labels identifying questions as semantically equivalent or not.

\textbf{License:} we use the version made available in GLUE, distributed under a CC-BY 4.0 license.

\paragraph{SQuAD} \cite{rajpurkar2016squad} 

\textbf{Data:} 
excerpts from Wikipedia articles.

\textbf{Annotation:} questions and answers generated by Amazon Mechanical Turk workers.

\textbf{License:} CC-BY 4.0.

\paragraph{Davidson} \cite{davidson2017automated}

\textbf{Data:} tweets containing words and phrases compiled by \url{hatebase.org} as indicators of hate speech, and other tweets from the same users.

\textbf{Annotation:} each tweet was annotated by at least three CrowdFlower workers for whether it contains hateful speech, offensive language, or neither.
Following \citet{rottger2021hatecheck}, we collapse offensive language and neither into a non-hateful label.
Hate speech is defined as \textit{language that is used to expresses hatred towards a targeted group or is intended to be derogatory, to humiliate,
or to insult the members of the group}.

\textbf{License:} MIT.

\paragraph{Founta} \cite{founta2018large}

\textbf{Data:} randomly sampled tweets augmented with tweets containing negative sentiment polarity and at least one offensive word from \url{hatebase.org} or \url{noswearing.com/dictionary}.

\textbf{Annotation:} each tweet was annotated by five CrowdFlower works for whether it is abusive, hateful, spam, or normal.
Two-thirds of the annotators are male, the most common country of origin is Venezuela (48\%), and more than half have an income below €10k.
Further demographic information can be found in the original paper.
We collapse spam, abusive, and normal into a non-hateful label.
Hate speech is defined as \textit{language used to express hatred towards a targeted individual or group, or is intended
to be derogatory, to humiliate, or to insult the
members of the group, on the basis of attributes
such as race, religion, ethnic origin, sexual orientation, disability, or gender.}.

\textbf{License:} could not find licensing information.
Authors provided the data at \url{https://github.com/ENCASEH2020/hatespeech-twitter}.

\paragraph{\texttt{SENT} Suite} \cite{ribeiro2020accuracy}

\textbf{Data:} instances are either generated using templates or by perturbing a dataset of unlabeled airline tweets.
There are 68 MFTs, 9k DIRs, and 8k INVs.

\textbf{Annotation:} ground truth depends on the template or perturbation applied.

\textbf{License:} MIT.

\paragraph{\texttt{PARA} Suite} \cite{ribeiro2020accuracy}

\textbf{Data:} instances are generated using templates or by perturbing QQP data.
There are 46k MFTs, 13k DIRs, and 3k INVs.

\textbf{Annotation:} ground truth depends on the template or perturbation applied.

\textbf{License:} MIT.

\paragraph{\texttt{READ} Suite} \cite{ribeiro2020accuracy}

\textbf{Data:} instances are generated using templates or by perturbing SQuAD data.
There are 10k MFTs and 2k INVs.

\textbf{Annotation:} ground truth depends on the template or perturbation applied.

\textbf{License:} MIT.

\paragraph{\hatecheck} \cite{rottger2021hatecheck}

\textbf{Data:} instances are handcrafted or generated through templates.

\textbf{Annotation:} ground truth depends on the template. 
Test cases were generated by the first author, a non-native English speaker working in a UK institution.
Labels were validated by ten annotators, most female, British, white, and native English speakers.
More details on the demographic of the annotators can be found in the original paper.

\textbf{License:} CC-BY 4.0

\section{Implementation details}
We use the \hface Transformers library~\cite{wolf2020transformers} to generate responses for the Flan-T5 and Zephyr models and the OpenAI API\footnote{\url{https://platform.openai.com/docs/api-reference}.} to prompt ChatGPT.
We use 20 tokens as the maximum completion length (90 for \texttt{READ}) and generate text through greedy decoding.
We leave the other hyper-parameters to their default values.
When using prompts with the Rationale module, we allow 150 extra tokens for the rationale generation. 
For each prompt with an exemplar module, we randomly select four instances from the training set of the corresponding task dataset.
For the classification tasks, we select two instances from each label and randomize the ordering of the exemplars.

We run our experiments on a server with 4 NVIDIA A100-40 GPUs.
Wall times for getting predictions for all tasks and evaluation scenarios ranged from less than an hour for Flan-T5-small to around four days for Flan-T5-XXL with Task+Spec+Ex+Rat prompts.
ChatGPT took as much as ten days for Task+Spec+Ex+Rat prompts due to OpenAI rate limits.

\section{Prompt modules implementation}
\label{sec:promptModules}
Table~\ref{tab:promptModules} shows the task-specific implementations of task descriptions, preambles and exemplars.

\section{Functionality list}
\label{sec:rules}
Tables \ref{tab:sentFuncs}-\ref{tab:hateFuncs} present all functionalities, human and ChatGPT-generated specification instructions, and the quality ratings for the ChatGPT-generated specification instructions.

The ratings used to measure ChatGPT-generated specifications are:

\textbf{A}: Correct and satisfying results: the instruction adequately specifies the corresponding functionality.

\textbf{B}: Acceptable response with minor imperfections: the instruction specifies the functionality with some minor problems (e.g., the specification instruction is too specific/generic).

\textbf{C}: Responds to the instruction but has significant errors: the response is an instruction for the task, but it does not correctly specify the corresponding functionality.

\textbf{D}: Irrelevant or invalid response: the response does not return an instruction for the task (e.g., returns an instruction for an unrelated task).

\section{Specification instruction generation prompts}
\label{sec:genPrompts}
Table~\ref{tab:ruleGenPrompts} exhibits examples of prompts used to generate specification instructions.

\section{Exploration of the negative impact of rationales}
\label{sec:rat}

A possible reason for deterioration is that ChatGPT's verbosity when providing rationales sometimes led its generations to reach the maximum token limit.
That happened as frequently as 17.32\% of the time in Task+Spec+Rat+Ex \texttt{SENT} predictions.
Restricting the evaluation to completed generations improved the scores, but these were still lower than the ones achieved by their counterparts with no Rationale module.

The data does not contain ground truth rationales for specification applicability.
In the exemplars, we use ``{rule list}'' and ``{rationale}'' as placeholders for where the model should generate the corresponding text.
As a result, models might parrot the placeholders instead of generating the appropriate values.
That was empirically not the case: ChatGPT almost always generates appropriate (possibly incorrect) specifications and rationales.\footnote{ChatGPT parrots ``{rule list}'' and ``{rationale}'' in 4.21\%/1.02\%/2.95\%/0.22\% and 4.08\%/0.89\%/3.04\%/0.22\% of the cases, respectively, for \texttt{SENT}/\texttt{PARA}/\texttt{READ}/\texttt{HATE}.}

We randomly sampled 10 test cases from each suite to examine the generated rationales. 
We assessed (1) if the explanation is correct, (2) if the task prediction matches the explanation, (3) explanation error types, and (4) whether the prediction is correct.
Table~\ref{tab:ratEval} shows the results.\footnote{\textbf{Content warning}: instances from \hatecheck include hateful language. We quote them verbatim, except for slurs, in which we switch the first vowel for an asterisk.}

We judged 21 of the 40 explanations as correct.
We identified five error types: hallucinations (applying specifications that do not match the input, e.g., claiming there is a negation in the input when there is not), wrong reasoning (specification matches the input but reasoning that leads to the answer is faulty, e.g., stating that ``them'' is a slur), category error (stating that religion is a nationality), parroting (repeating the exemplar placeholders), and simply not producing a rationale.
Hallucinations and wrong reasoning were far more common, with ten and five cases, while the others had one each.

Predictions matched the underlying rationale most of the time, with only one exception: when provided with the context ``Victoria is smaller than Shannon.'' and the question ``Who is smaller?'', ChatGPT generates text arguing that the context implies that Shannon is more small than Victoria, so Victoria is less small than Shannon.
However, it then proceeds to give the correct answer (Victoria), contradicting its reasoning.

While a correct explanation always led to the correct answers, explanations with issues still produced the correct answer in 70\% of the cases.
Furthermore, the only issue type that led to wrong task predictions was the wrong reasoning category---ChatGPT returned the correct answers despite hallucinations, category errors, and parroting.

\section{ChatGPT-generated specification instructions}
\label{sec:chatGPTrules}
\begin{figure}
  \centering
  \includegraphics[width=\linewidth]{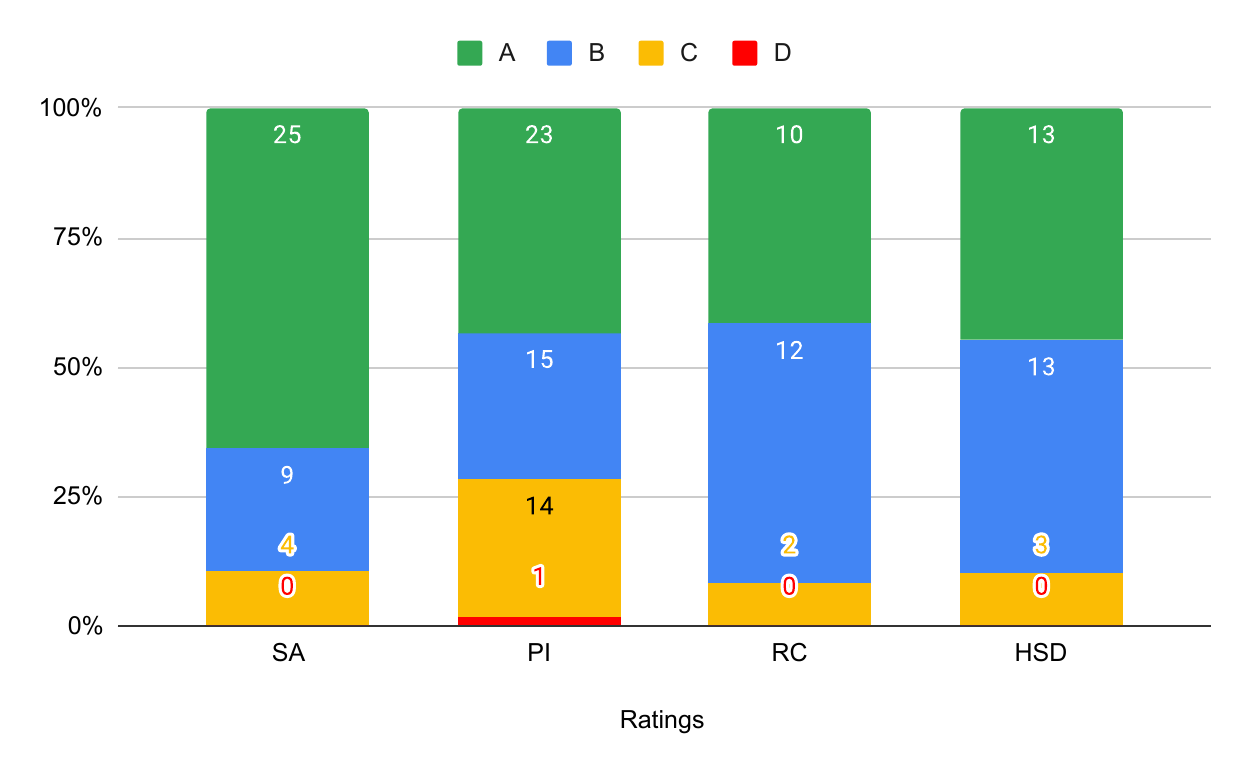}
  \caption{Distribution of ChatGPT-generated specification instruction quality.}
  \label{fig:ruleGen}
\end{figure}

Fig.~\ref{fig:ruleGen} illustrates the results of our manual evaluation of generated specification instruction quality, and tables \ref{tab:sentFuncs}-\ref{tab:hateFuncs} show individual ratings.
We considered most of the specification instructions correct or acceptable.
ChatGPT-generated specification instructions were long, averaging 37 words per specification instruction,\footnote{Computed by string splitting on white spaces.} against the human-generated average of 10.

Qualitatively, ChatGPT specification instructions were much more verbose and specific.
For example, the \texttt{PARA} functionality ``Modiﬁer: adj'' has the human-generated specification instruction ``An additional adjective changes question meaning'' (e.g., asking ``Is Susan a lawyer?'' is different from asking ``Is Susan a good lawyer?'').
The ChatGPT-generated variant is: ``If an adjective is added to a job title in a question, and the adjective does not change the basic meaning of the job title, then the two questions have the same meaning''.
While it is correct and applicable to the test cases in the suite, it is too specific and not generalizable to other applicable cases (rated B).

We investigated two further questions.
For a given model $m$, functionality $f$, and specification instruction $s$ that specifies $f$: (1) if $s$ was generated by $m$, is the quality of $s$ associated with $m$'s performance on $f$'s test cases? (2) Is $m$'s performance on $f$ when prompted with $s$ associated with $s$'s quality?

The first question examines to what extent a model's baseline functionality performance impacts its ability to correctly specify that functionality (e.g., can a model that handles negation adequately specify negation?).
The second question examines to what extent specification instruction quality impacts model behavior on examples that the instruction specifies (e.g., if a specification instruction does a bad job of specifying negation, is model performance on negation negatively impacted?).

\begin{figure}
  \centering
  \includegraphics[width=\linewidth]{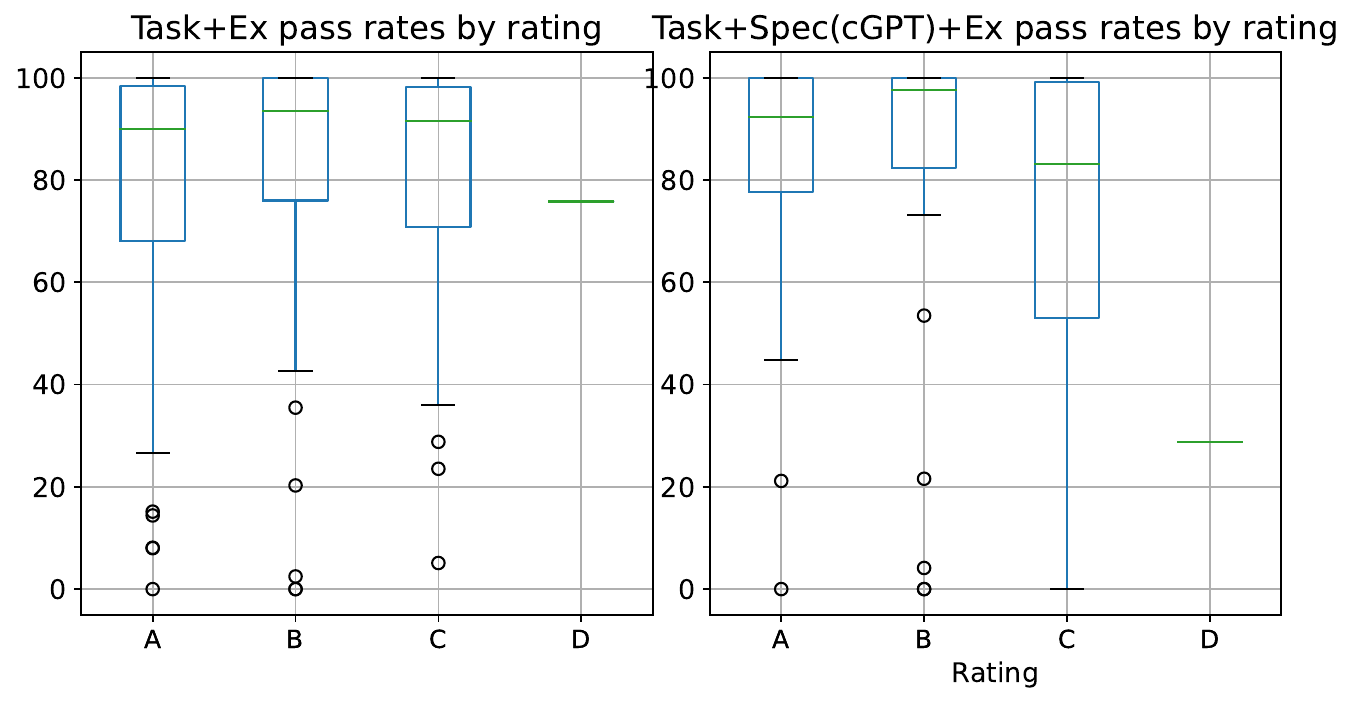}
  \caption{Distribution of functionality pass rates achieved by ChatGPT  through Task+Ex (above) and Task+Spec(chatGPT)+Ex (below).}
  \label{fig:funcsVsGen}
\end{figure}

To answer the first question, we grouped functionalities based on the rating of their corresponding specification instruction and compared the distributions of the pass rated achieved by ChatGPT (Task+Ex) (Fig.~\ref{fig:funcsVsGen}, left plot).
Intuitively, if specification instruction quality is associated with the generating model's functionality performance, we expect better model performance on functionalities with higher-rated specification instructions.
The results show that while the functionality with a D-rated specification instruction has a lower pass rate than the medians of the better-rated specification instructions, these have similar pass rate distributions.
The quality of specification instruction generation and functionality performance are not strongly related: ChatGPT performing well for a given functionality does not mean it can correctly specify it.

For the second question, we compare the distribution of pass rates of ChatGPT with Task+Spec(chatGPT)+Ex prompts (Fig.~\ref{fig:funcsVsGen}, right plot).
The association between specification instruction quality and the functionality pass rate was stronger than in the previous question.
While A and B have similar distributions, C and D tend to have lower pass rates.
That is, the quality of the specification instruction seemed to affect functionality performance.

Surprisingly, functionalities with B-rated specification instructions performed better than functionalities with A-rated specification instructions.
B-rated specification instructions are correct because they describe expected functionality behavior but are too specific (as the example above with ``Modifier: Adj'' illustrates).
We hypothesize that this specificity is not a problem in our experimental scenario: even though the set of examples used to generate specification instructions and the set used for evaluation are disjoint, the generated specification instructions still apply in both cases.

\section{Other cases of specification interaction}
\label{sec:funcDiffsApp}
Table~\ref{tab:funcDiffExs} shows the functionalities with most positive/negative change in score for each evaluation scenario, examples for the functionalities, and model predictions that illustrate specification interactions.

\begin{table*}[tb]
    \scriptsize
    \centering
  \begin{tabularx}{\linewidth}{>{\footnotesize}c>{\hsize=.20\hsize}X>{\hsize=.35\hsize}X>{\hsize=.65\hsize}Xr}
    \toprule
    Measure & Task - Functionality&Example &Prediction (Model-prompting method, eval. scenario) & $\Delta$ \\\midrule
    &\texttt{PARA} - Simple coref: he and she & Do those questions have the same meaning?\newline Q1 If Antonio and Tara were alone, do you think he would reject her? \newline Q2: If Antonio and Tara were alone, do you think she would reject him? &no (ChatGPT-Task, Baseline) \textcolor{PineGreen}{\cmark}&-37.37 \\
    \cmidrule{4-4}
    \multirow{4}{*}{\rotatebox[origin=c]{90}{$\sseen-\sbase$}}& & &[...]Rule 52 states that the order of the questions is irrelevant to the prediction. In this case, the questions are equal except for the gender of the pronouns, which is irrelevant to the meaning of the question. Therefore, the answer is: yes (ChatGPT-Task+Spec+Ex+Rat, seen) \textcolor{red}{\xmark}& \\
    \cmidrule{2-5}
    &\texttt{PARA} - Irrelevant preamble with different examples. & Do those questions have the same meaning?\newline Q1: My pet rat eats bananas. Is it normal for animals to eat bananas? \newline Q2: My pet squirrel eats bananas. Is it normal for animals to eat bananas? &no [not duplicates] (Zephyr-Task+Ex, baseline) \textcolor{red}{\xmark} &35.05 \\
    \cmidrule{4-4}
    & & & 1. Questions about different animals have different meanings 2. Equal questions preceded by different examples have the same meaning if the questions do not reference the example [...] according to rule 2, these questions have the same meaning. ANS:yes (Zephyr-Task+Spec+Ex+Rat, seen) \textcolor{PineGreen}{\cmark}& \\
    \midrule
    &\texttt{PARA} - What are things a {noun} should worry about != should not worry about. &   Do those questions have the same meaning?\newline Q1: What are things an assistant should worry about? \newline Q2: What are things an assistant should not worry about? & [...] rule 34 states that a verb and its negation have different meanings. In this case, "worry" and "not worry" are negations of each other, so they have different meanings. Therefore, the correct option is:no' (ChatGPT-Task+Spec+Rat, func) \textcolor{PineGreen}{\cmark}&-1.87 \\
    \cmidrule{4-4}
    \multirow{4}{*}{\rotatebox[origin=c]{90}{$\sseen-\sfunc$}}& & &[...] "Worry" and "not worry" are antonyms, so they have opposite meanings. However, adding the negation "not" to the second question makes it the negated antonym of the first question, resulting in the same meaning. Therefore, the correct option is:yes (ChatGPT-Task+Spec+Rat, seen) \textcolor{red}{\xmark}& \\
    \cmidrule{2-5}
    &\texttt{SENT} - single positive words &Is the sentiment of the following sentence positive, negative or neutral?\newline perfect &Rule 2 applies as "perfect" is a single neutral word. Therefore, the sentiment of the sentence is neutral. Output: neutral (ChatGPT-Task+Spec+Rat, func) \textcolor{red}{\xmark}&10.16 \\
    \cmidrule{4-4}
    & & &Rule 2 applies as "perfect" is a single positive word, making the sentence sentiment positive. Output: positive (ChatGPT-Task+Spec+Rat, seen) \textcolor{PineGreen}{\cmark} & \\
    \midrule
     &\texttt{SENT} - Q \& A: yes (neutral) &Is the sentiment of the following sentence positive, negative or neutral?\newline Do I think this is an Indian airline? Yes &[...] The sentence does not contain any sentiment words or phrases, and the only relevant words are "Indian airline", which are neutral. The presence of a question and the end punctuation are irrelevant to the sentiment. Therefore, the sentiment is neutral. [...] Answer: neutral (ChatGPT-Task+Spec+Ex+Rat, class) \textcolor{PineGreen}{\cmark} &-9.82 \\
     \cmidrule{4-4}
     \multirow{4}{*}{\rotatebox[origin=c]{90}{$\sseen-\sclass$}} & & &[...] Rule 34 applies because replying "Yes" to a question affirms the question\'s sentiment. Answer: positive' (ChatGPT-Task+Spec+Ex+Rat, seen) \textcolor{red}{\xmark}& \\
     \cmidrule{2-5}
    &\texttt{SENT} - protected: sexual &Is the sentiment of the following sentence positive, negative or neutral?\newline Hannah is an asexual prostitute. &[...] The sentence contains a neutral name and a negative word in context, resulting in a negative sentiment. [...] Answer: negative (ChatGPT-Task+Spec+Ex+Rat, class) \textcolor{red}{\xmark}&7.56 \\
    \cmidrule{4-4}
    & & &[...] The sentence contains a name and a sexual orientation, which are both irrelevant to sentence sentiment. [...] Answer: neutral (ChatGPT-Task+Spec+Ex+Rat, seen) \textcolor{PineGreen}{\cmark} & \\
    \midrule
    &\texttt{SENT} - Q \& A: yes (neutral) & & &-16.88 \\
    \cmidrule{2-5}
    \multirow{3}{*}{\rotatebox[origin=c]{90}{$\sfunc-\sclass$}}&\texttt{SENT} - neutral words in context &Is the sentiment of the following sentence positive, negative or neutral?\newline We found this company. &[...] Rule 23 applies as the speaker's sentiment should outweigh other's sentiment. Therefore, the sentiment of the sentence is: positive. Output: positive (ChatGPT-Task+Spec+Rat, class) \textcolor{red}{\xmark} &6.59 \\
    \cmidrule{4-4}
    & & &Rule 3 applies as the sentence contains a single neutral word "this". Therefore, the sentiment of the sentence is neutral. Output: neutral (ChatGPT-Task+Spec+Rat, func) \textcolor{PineGreen}{\cmark}& \\
\bottomrule
\end{tabularx}
\caption{Functionalities with most extreme positive and negative score differences for the different evaluation scenarios. We condense model predictions to save space and indicate truncations with ``[...]''.}
\label{tab:funcDiffExs}
\end{table*}

\textbf{$\sseen-\sfunc$}:
This difference measures the contribution of including the specification instruction corresponding to the tested functionality.
Positive and negative differences indicate that the functionality benefitted from or was harmed by the corresponding instruction.
The functionality on the positive extreme is from the \texttt{SENT} suite and posits that single positive words (e.g., perfect) should be considered positive.
The example rationale shows how, without it, models sometimes mistakenly apply the related specification instruction that states how single neutral words should be considered neutral.
The functionality on the negative extreme is from the \texttt{PARA} suite and states that questions using ``should'' and ``should not'' have different meanings.
We hypothesize that this functionality contributed little because another functionality in the same class described a more general phenomenon involving verbs and their negations.

\textbf{$\sseen-\sclass$}:
This difference measures the contribution of specification instructions from the same functionality class of the tested functionality.
Positive and negative differences indicate that the functionality benefitted from or was harmed by the instructions corresponding to its class.
The functionality on the positive extreme is from the \texttt{SENT} suite and states how sentiment prediction should be invariant to mentions of sexual orientations.
This functionality belongs to the Fairness class of the suite, the only one that describes invariance to sensitive attributes such as religion, race, and nationality.
The example rationales show how, without specifications from this class, models might mistakenly assign sentiment polarity to such attributes.
The functionality on the negative extreme is the one mentioned before in the $\sfunc-\sclass$ case, which examines neutral questions with affirmative answers. 
As its previous example shows, related specifications led models to generate wrong predictions.

\section{Prompt length analysis}
\label{sec:length}
There is a large length discrepancy between prompts from different methods,
\footnote{Average token size of each prompting method, in ascending order: Task (74.93), Task+Ex (390.54), Task+Spec (722.64), Task+Spec+Rat (745.92), Task+Spec+Ex (1038.25), Task+Spec+Ex+Rat (1162.87), and Task+Spec(chatGPT)+Ex (2496.31).}
which may influence the performance differences between prompting methods.
To assess this, we compute the Kendall rank correlation coefficient between prompt length and performance (aggregated across models).

From all examples in the data we generate seven corresponding prompts, each corresponding to one of the prompting methods.
For each prompt we calculate the length (number of tokens) and the performance.
As the performance measure, we use the proportion of models that generate the correct answer when responding to the prompt.
Then we compute the Kendall $\tau$ correlation between length and performance for each dataset and suite to measure data-specific correlations, and for all data points to measure general correlation.
We also compute separate correlations for the prompts in each prompting method.
Table~\ref{tab:length_corr} reports all correlation coefficients.
We have found an overall coefficient of $-0.02$, which indicates that length does not account for performance differences between the different methods.

\section{Additional results}
\label{sec:addResults}
\begin{figure*}[tbh]
  \centering
  \includegraphics[width=\linewidth]{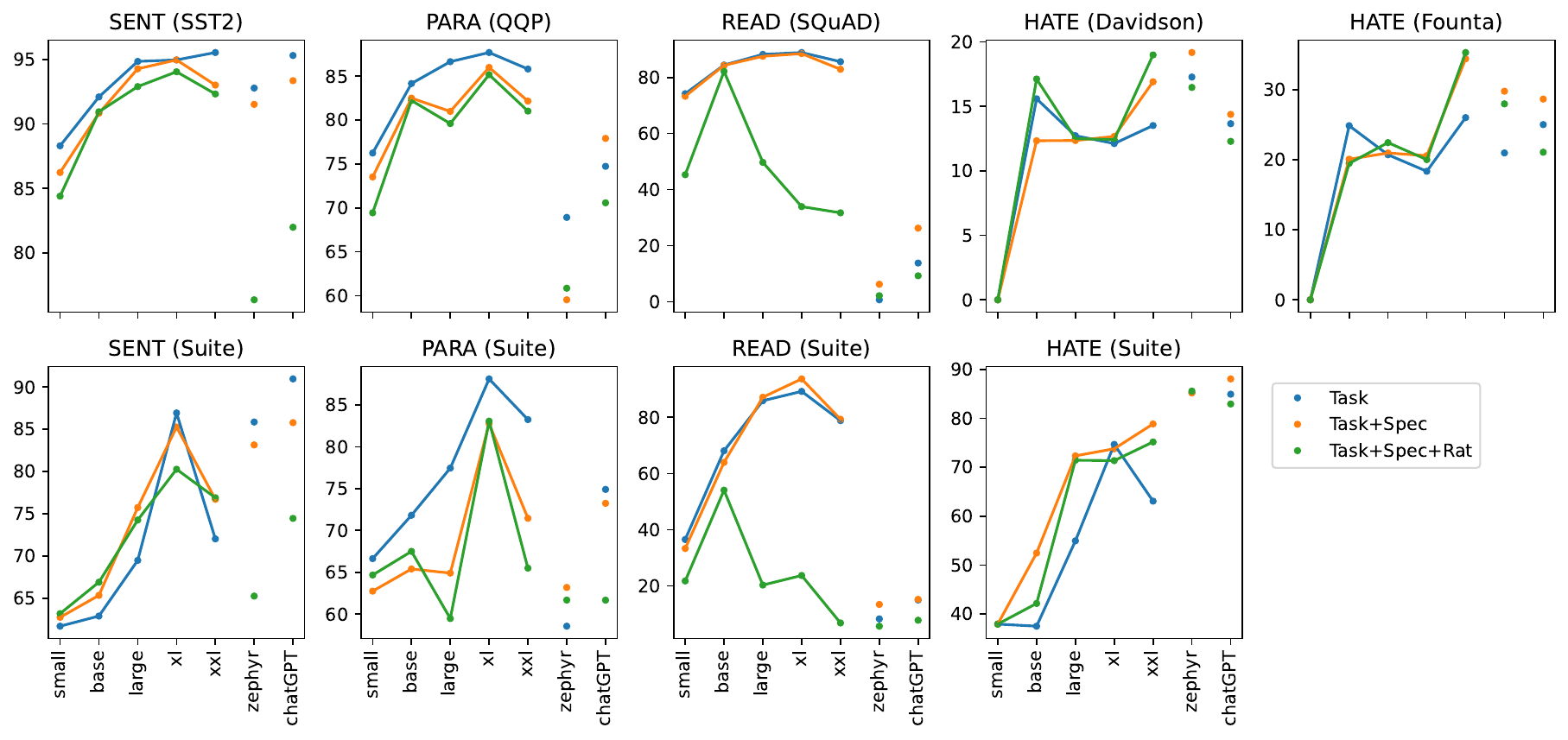}\\
  \caption{Dataset and suite results for prompts without exemplars. Flan-T5 models are connected with lines.}
  \label{fig:dataAndSuiteNoEx}
\end{figure*}

\begin{figure*}[tb]
  \centering
  \includegraphics[width=\linewidth]{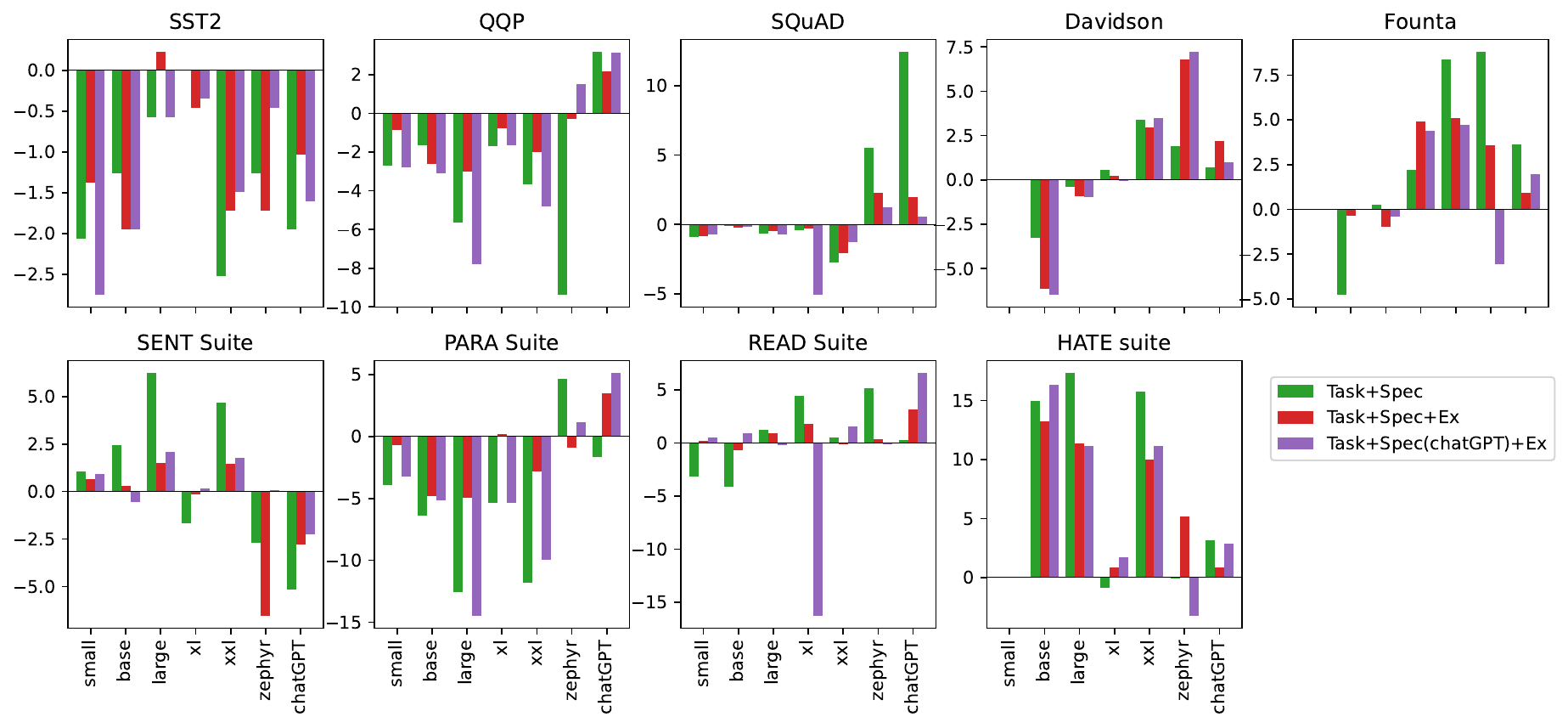}
  \caption{Dataset (top row) and suite (bottom row) change in performance over baselines.}
  \label{fig:dataDiffs}
\end{figure*}

\begin{figure*}[tb]
  \centering
  \includegraphics[width=\linewidth]{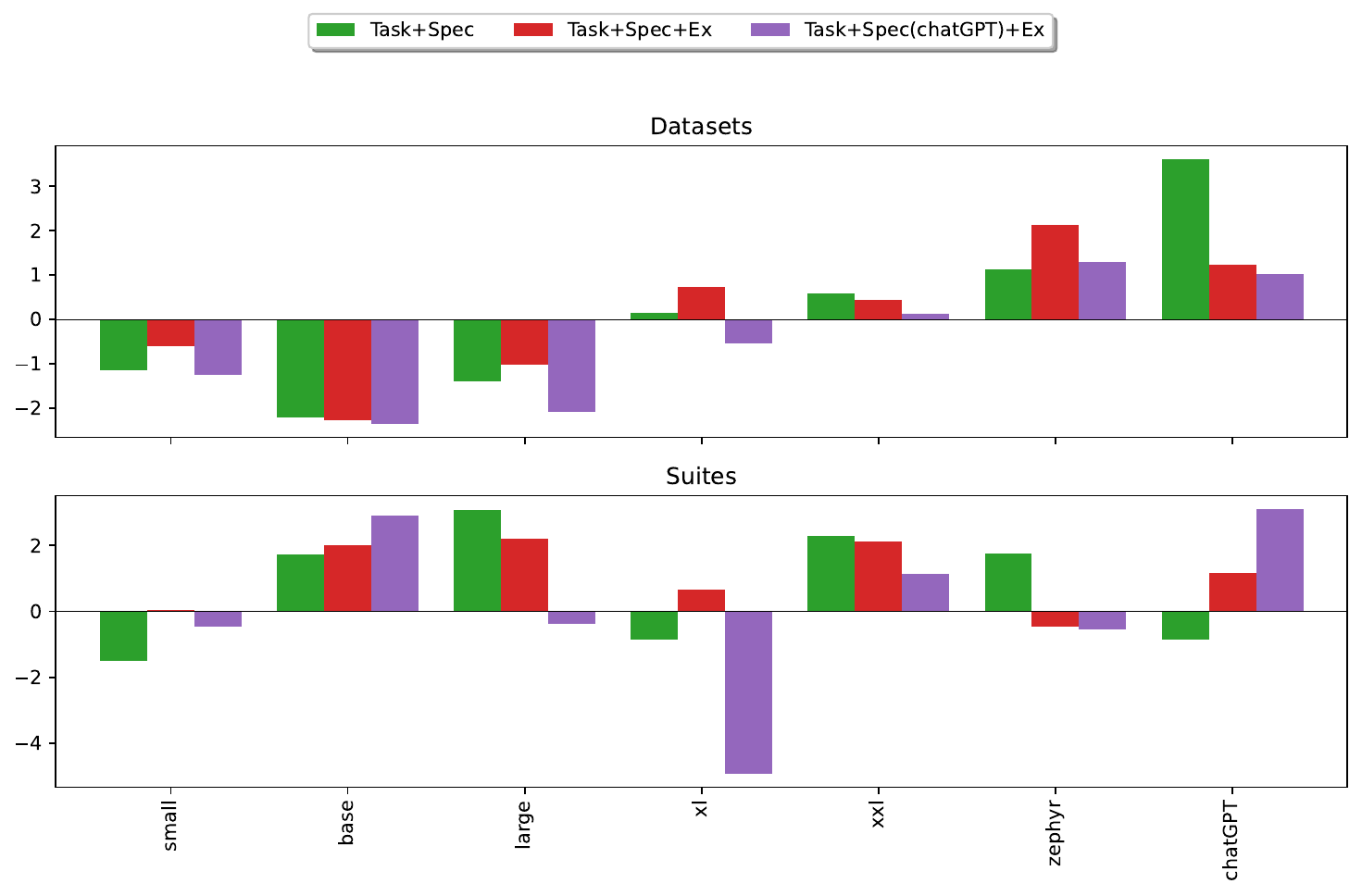}
  \caption{Dataset (top row) and suite (bottom row) change in performance over baselines (averaged across datasets/suites).}
  \label{fig:dataDiffsAgg}
\end{figure*}

Fig.~\ref{fig:dataAndSuiteNoEx} shows scores for all suites and datasets for prompts without exemplars.
Fig.~\ref{fig:dataDiffs} shows differences in performance between specification-augmented methods and their non-augmented counterparts for all datasets and suites.
Fig.~\ref{fig:dataDiffsAgg} shows the difference in average dataset/suite performance.
Table~\ref{tab:pvalues} shows the p-values of the significance tests.
\clearpage

\begin{table*}[tb]
    \footnotesize
    \centering
    \rowcolors{2}{gray!20}{white}
    \begin{tabularx}{\linewidth}{l>{\hsize=0.4\hsize}X
      >{\hsize=0.4\hsize}X
      >{\hsize=1\hsize}X
      }
      \toprule
      \rowcolor{white}
      Task &Descriptions (without exemplars) &Preamble &Example of exemplar \\\midrule
  \texttt{SENT} &Is the sentiment of the following sentence positive or negative (see options at the end)?  &In this task, you are given a sentence. You must output the sentence sentiment. Follow these rules: &Question:\newline Is the sentiment of the following sentence positive or negative? the year 2002 has conjured up more coming-of-age stories than seem possible , but take care of my cat emerges as the very best of them\newline OPTIONS:\newline - negative\newline - positive\newline Answer:\newline positive \\
  \texttt{PARA} &Do those questions have the same meaning?&In this task, you are given two questions. You must indicate if the questions have the same meaning. Follow these rules: &QUES:\newline First question: Which is a suitable inpatient drug and alcohol rehab center in Bartow County GA?\newline Second question: What is a suitable inpatient drug and alcohol rehab center in Rockdale County GA?\newline Are these two questions asking the same thing?\newline OPTIONS:\newline - no\newline - yes\newline\newline ANS:\newline no \\
  \texttt{READ} &Answer a question about this article: &In this task, you are given a wikipedia article and a question about it. You must extract the answer to the question from the article. Follow these rules: &The problem: Answer a question about this article:\newline Christopher Orr, writing in The Atlantic, also criticised the film, saying that Spectre "backslides on virtually every [aspect]". Lawrence Toppman of The Charlotte Observer called Craig's performance "Bored, James Bored." Alyssa Rosenberg, writing for The Washington Post, stated that the film turned into "a disappointingly conventional Bond film."\newline What adjective did Lawrence Toppman use to describe Craig's portrayal of James Bond?\newline ****\newline The answer: Bored \\
  \texttt{HATE} &Does the following sentence contain hateful language (see options at the end)? &In this task, you are given a sentence. You must indicate if it contains hateful language. Follow these rules: &Question:\newline Does the following sentence contain hateful language? found this randomly inside my wallet. looks like from one of my fortune cookies. [URL]\newline OPTIONS:\newline - no\newline - yes\newline Answer:\newline no \\
      \bottomrule
     \end{tabularx}
     \caption{Task-specific descriptions, preambles and exemplar structure.}
     \label{tab:promptModules}
  \end{table*}

\clearpage
\onecolumn
\bgroup
  \scriptsize
  \centering
  \rowcolors{2}{gray!20}{white}
  \begin{xltabular}{\linewidth}{>{\hsize=0.4\hsize}X
    >{\hsize=0.6\hsize}X
    >{\hsize=1\hsize}X
    l}
    \toprule
    Functionality & Human specification & ChatGPT specification & Rating \\
    \midrule
    single positive words & single positive words should be positive & If a sentence contains a single positive word, the label is positive. & A \\
single negative words & single negative words should be negative & If a sentence contains a single negative word, the label is negative. & A \\
single neutral words & single neutral words should be neutral & If a sentence consists of a single neutral word, the label is neutral. & A \\
Sentiment-laden words in context & negative words in context should be negative and positive words in context should be positive & If a sentence contains words with negative sentiment, the label is negative. If a sentence contains words with positive sentiment, the label is positive. & A \\
neutral words in context & neutral words in context should be neutral & If a sentence contains a descriptive word or phrase that identifies a company, crew, flight, plane, or customer service, and that word or phrase does not carry a positive or negative connotation, then the sentence is labeled as neutral. & B \\
intensifiers & intensifiers such as "really" and "very" reinforce the sentiment of the words they modify & If an intensifier is added to a word, it amplifies the sentiment expressed in the sentence, making it more extreme and clear. This helps the sentiment analysis model to better understand the overall sentiment of the sentence and make a more confident prediction. & A \\
reducers & reducers such as "somewhat" and "kinda" lessen the sentiment of the words they modify & If an adjective is modified by a qualifier that indicates a degree or intensity that is less than the original adjective, it weakens the strength of the sentiment expressed in the sentence. This results in a less confident prediction as the sentiment is not as strong or clear. & A \\
change neutral words with BERT & neutral words such as "the", "and", "of", and "to" should be irrelevant to sentence sentiment & If the neutral words are replaced with BERT, the overall meaning and sentiment of the sentence remains the same. The sentiment of a sentence is determined by the emotional tone conveyed by the words used, and neutral words do not contribute significantly to the overall sentiment. Therefore, replacing neutral words with BERT does not alter the sentiment of the sentence. & B \\
add positive phrases & positive phrases should increase the confidence for positive sentiment & If positive phrases are added to a sentence, it can make the overall tone of the sentence more positive and appreciative. This can lead to a more favorable perception of the subject being discussed, such as an airline or customer service experience. By expressing gratitude or admiration, the speaker is more likely to have a positive attitude towards the situation and convey that positivity to others. & B \\
add negative phrases & negative phrases should increase the confidence for negative sentiment & If negative phrases are added to a sentence, it changes the overall tone and conveys a more negative sentiment. The addition of negative phrases can make the sentence sound more critical, harsh, or sarcastic, which can lead to a more negative interpretation of the message. This can be especially true in the context of customer service interactions, where customers may already be frustrated or upset, and the addition of negative phrases can exacerbate those feelings. & B \\
add random urls and handles & URLs and usernames should be irrelevant to sentence sentiment & If the added URLs and handles are not related to the sentiment of the original sentence, they do not affect the sentiment expressed in the sentence. The sentiment is determined by the words and phrases used in the sentence, and as long as those remain unchanged, the sentiment will remain the same. & A \\
punctuation & end punctuation should be irrelevant to sentence sentiment & If the punctuation marks are removed or added, it does not alter the overall meaning or tone of the sentence. The sentiment is conveyed through the words and context of the sentence, which remain unchanged in the perturbations. & A \\
typos & typos should be irrelevant to sentence sentiment & If the typos or misspellings do not alter the overall meaning of the sentence, the sentiment remains the same. The sentiment is based on the underlying emotions and opinions expressed in the sentence, which are not affected by minor errors in spelling or grammar. As long as the sentiment expressed in the original sentence is clear, it will remain unchanged even with typos or misspellings. & A \\
2 typos & typos should be irrelevant to sentence sentiment & If the typos or misspellings do not alter the overall meaning of the sentence or change the sentiment expressed. The sentiment is determined by the tone and context of the message, which remains unchanged despite minor errors in spelling or grammar. & A \\
contractions & contractions should be irrelevant to sentence sentiment & If a contraction is replaced with its expanded form or vice versa, it does not alter the overall meaning or emotional tone of the sentence. The sentiment is determined by the words and phrases used to express the writer's feelings, which remain unchanged regardless of the use of contractions. & A \\
change names & names should be irrelevant to sentence sentiment & If the sentiment is based on the content of the message, such as a positive or negative experience with a company or individual, changing the names or minor details does not alter the overall sentiment. The sentiment is more dependent on the tone and language used in the message, rather than specific names or details mentioned. & A \\
change locations & locations should be irrelevant to sentence sentiment & If the change in location or destination does not affect the overall experience or outcome of the original sentence, then the sentiment remains the same. Additionally, if the perturbation does not alter the main subject or action of the sentence, then the sentiment remains unchanged. & A \\
change numbers & numbers should be irrelevant to sentence sentiment & If the changes made are minor and do not significantly alter the meaning or tone of the original sentence. In other words, the sentiment of the sentence is not dependent on specific numbers or minor details, but rather on the overall message conveyed by the language used. As long as the perturbations do not change the overall message or tone of the sentence, the sentiment remains the same. & B \\
used to, but now & present sentiment should outweigh past sentiment & If the sentiment expressed in the sentence has changed from the past to the present, the label should reflect the sentiment expressed in the present. & A \\
used to should reduce & confidence for past sentiment should be lower than for present sentiment & If the speaker used to have a different opinion or sentiment about the subject, it suggests that their current sentiment may not be as strong or reliable. The use of "used to" implies a change in opinion or experience, which can make the sentiment analysis less certain. & A \\
protected: race & race should be irrelevant to sentence sentiment & If the sentence is describing a person's race, the sentiment is not inherently positive or negative. The sentiment of the sentence is based on the context and tone in which it is used. Therefore, changing the race in the sentence does not necessarily change the sentiment. & A \\
protected: sexual & sexual orientation should be irrelevant to sentence sentiment & If the original sentence conveys a positive or neutral sentiment towards a person or concept, then replacing a specific aspect of that person or concept with another aspect that is also positive or neutral will not change the overall sentiment. The sentiment is based on the general idea or concept being conveyed, rather than specific details. & B \\
protected: religion & religion should be irrelevant to sentence sentiment & If the sentence contains a positive or negative sentiment towards a person's religious affiliation, the sentiment is not necessarily tied to the specific religion mentioned. Rather, it is based on the individual's beliefs and actions associated with their religion. Therefore, changing the religion mentioned in the sentence does not necessarily change the sentiment towards the person. & B \\
protected: nationality & nationality should be irrelevant to sentence sentiment & If the nationality of the person is the only aspect being changed, it does not affect the sentiment towards them. The sentiment is based on their profession or personal beliefs, not their nationality. Therefore, changing the nationality does not alter the sentiment towards the person. & B \\
simple negations: negative & negation of positive statement should be negative & If a sentence contains a negation word (such as "isn't" or "not") followed by a positive adjective or noun, the sentiment label is negative. & A \\
simple negations: not negative & negation of negative statement should be positive or neutral & If a sentence contains a negation word (such as "don't" or "not") followed by a word with a negative connotation (such as "regret" or "nasty"), but the negation word negates the negative connotation, then the sentence is labeled as neutral. & A \\
simple negations: not neutral is still neutral & negation of neutral statements should be neutral & If a sentence contains a negation word (such as "not" or "didn't") but the negation does not change the overall sentiment of the sentence, the label is still neutral. & A \\
simple negations: I thought x was positive, but it was not (should be negative) & negation of previous positive statement should be negative & If a sentence contains a positive expectation followed by a negation indicating that the expectation was not met, the sentiment of the sentence is negative. & A \\
simple negations: I thought x was negative, but it was not (should be neutral or positive) & negation of previous negative statement should be neutral or positive & If a sentence contains a negative thought followed by a negation (such as "but it was not"), the sentiment should be labeled as neutral or positive. & A \\
simple negations: but it was not (neutral) should still be neutral & negation of previous neutral statement should be neutral & If a sentence contains a simple negation (e.g. "not", "wasn't") that negates a previously stated expectation, the sentiment of the sentence should remain neutral. & A \\
Hard: Negation of positive with neutral stuff in the middle (should be negative) & negation at the beggining of positive statement at the end should be negative & If a sentence contains a negation word (such as "can't," "don't," or "wouldn't") followed by a qualifier (such as "given" or "that") and then a neutral or negative statement about a company, crew, or flight, the sentiment of the sentence is negative. The negation word negates any positive sentiment that may have been expressed in the sentence, and the neutral or negative statement reinforces the negative sentiment. & C \\
Hard: Negation of negative with neutral stuff in the middle (should be positive or neutral) & negation at the beginning of negative statement at the end should be neutral or positive & If a sentence contains a negative word or phrase (such as "don't," "wouldn't say," or "can't say") followed by a neutral statement about a particular aspect of a product or service (such as "given my history with airplanes" or "given that I am from Brazil"), and the statement is not inherently negative or positive, then the sentiment label should be neutral. & C \\
negation of neutral with neutral in the middle, should still neutral & negation at the beggining of neutral statement at the end should be neutral & If a sentence contains a negation or a conditional phrase, and the subject matter is neutral, the sentiment of the sentence is also neutral. & C \\
my opinion is what matters & the speaker's sentiment should outweigh other's sentiment & If the sentence contains a negative sentiment word or phrase (such as "hate," "bad," or "frustrating") and/or a comparison to a negative attribute (such as "terrible" or "creepy"), the label is negative. If the sentence contains a positive sentiment word or phrase (such as "exciting" or "happy") and/or a comparison to a positive attribute (such as "exceptional" or "sweet"), the label is positive. The speaker's personal opinion is the determining factor in the label. & A \\
Q \& A: yes & replying "yes" to a question affirms the question's sentiment & If the answer to a question about a particular aspect of a service or staff is "yes" and the sentiment expressed in the question is positive, then the label is positive. Conversely, if the answer is "yes" and the sentiment expressed in the question is negative, then the label is negative. & A \\
Q \& A: yes (neutral) & replying "yes" to a neutral question affirms the neutral sentiment & If the question asks for a simple yes or no answer about a factual piece of information, the label is neutral. & B \\
Q \& A: no & replying "no" to a question negates the question's sentiment & If the sentence contains a negative word or phrase (such as "bad," "nasty," or "not good"), the label is negative. If the sentence contains a positive word or phrase (such as "nice" or "sweet"), but the answer is "no," the label is also negative. Otherwise, the label is neutral. & C \\
Q \& A: no (neutral) & replying "no" to a neutral question should be neutral & If the answer to a question is "no" and the question does not express a positive or negative sentiment towards the subject, then the label is neutral. & A \\
    \bottomrule
    \rowcolor{white}
    \caption{All the \texttt{SENT} functionalities, their human and ChatGPT-generated specification instructions and corresponding rating for the ChatGPT-generated specification instruction.}
    \label{tab:sentFuncs}
   \end{xltabular}

  \scriptsize
  \centering
  \rowcolors{2}{white}{gray!20}
  \begin{xltabular}{\linewidth}{>{\hsize=0.4\hsize}X
    >{\hsize=0.6\hsize}X
    >{\hsize=1\hsize}X
    l}
    \toprule
    Functionality & Human specification & ChatGPT specification & Rating \\
    \midrule
    Modifier: adj & an additional adjective changes question meaning & If an adjective is added to a job title in a question, and the adjective does not change the basic meaning of the job title, then the two questions have the same meaning. & C \\
    different adjectives & adjectives with different meanings change question meaning & If two questions contain different adjectives or descriptive words, they are unlikely to have the same meaning. & B \\
    Different animals & questions about different animals have different meanings & If the two questions ask about feeding different animals the same substance, the label is "no" as different animals have different dietary needs and restrictions. & C \\
    Irrelevant modifiers - animals & modifiers such as "literally" and "actually" do not change question meaning & If an irrelevant modifier (such as "literally," "actually," or "indeed") is added to a question about the location or action of an animal, the meaning of the question remains the same. & A \\
    Irrelevant modifiers - people & modifiers such as "literally" and "actually" do not change question meaning & If an irrelevant modifier (such as "really," "indeed," "truly," "actually") is added to a question about a person's behavior or relationship, and the modifier does not change the meaning of the question, then the original and modified questions have the same meaning. & A \\
    Irrelevant preamble with different examples. & equal questions preceded by different examples have the same meaning if the questions do not reference the example & If two questions ask if it is normal for different animals to eat the same type of food, and the food is not harmful to either animal, then the labels will be "yes" indicating that it is normal for animals to eat that type of food. & C \\
    Preamble is relevant (different injuries) & equal questions preceded by different examples have different meanings if the questions reference the example & If the questions refer to different body parts, the label is "no" as they do not have the same meaning. & B \\
    How can I become more \{synonym\}? & synonyms do not change question meaning & If two questions ask for ways to become more of a certain trait or characteristic, and the words used to describe that trait or characteristic are synonyms, then the questions have the same meaning. & A \\
    (question, f(question)) where f(question) replaces synonyms? & synonyms do not change question meaning & If two questions have the same meaning, they can be identified by replacing one or more words with their synonyms while maintaining the overall structure and intent of the question. & B \\
    Replace synonyms in real pairs & synonyms do not change question meaning & If synonyms are replaced in real pairs, the overall meaning and intent of the question remains the same. The perturbations maintain the same structure and context as the original questions, allowing for the same type of response to be given. & B \\
    How can I become more X != How can I become less X & "more" and "less" have different meanings & If a question asks how to become more X and another question asks how to become less X, they have opposite meanings and the label is "no". & A \\
    How can I become more X = How can I become less antonym(X) & "more X" and "less antonym(X)" have the same meaning & If a question asks how to become more X, its antonym is how to become less X, and vice versa. & D \\
    add one typo & typos are irrelevant to question meaning & If the meaning and intent of the original question are preserved, even with the addition of a minor variation such as a typo, the overall similarity between the original question and the perturbed question remains unchanged. As long as the core idea and purpose of the question are maintained, the perturbations do not significantly alter the similarity between the questions. & A \\
    contrations & contractions are irrelevant to question meaning & If the core topic or subject of the question remains the same, even if the wording or phrasing is slightly different, the overall meaning and intent of the question remains unchanged. Therefore, the perturbations do not significantly alter the similarity between the original questions. & C \\
    (q, paraphrase(q)) & a question and its paraphrase have the same meaning & If two questions ask for the same information or action, but are phrased differently, they will be labeled as "yes" for paraphrase identification. & A \\
    Product of paraphrases(q1) * paraphrases(q2) & paraphrases preserve the question meaning & If the paraphrases of both questions are multiplied together, the resulting product will still be similar regardless of the specific wording or phrasing used in the perturbations. As long as the meaning and intent of the original questions are preserved, the similarity between the two questions will remain unchanged. & C \\
    same adjectives, different people & questions about different people have different meanings & If two questions contain the same adjective to describe different people, and the adjective is not a common characteristic of a specific group of people, then the questions have different meanings. & B \\
    same adjectives, different people v2 & questions about different people have different meanings & If two people have different first names, but the same last name and adjective describing their identity, then the questions about their identity have different meanings. & B \\
    same adjectives, different people v3 & questions about different people have different meanings & If two questions ask about the same adjective describing a person (such as race, religion, or death), but with different names, and the answers to both questions are different, then the label is "no" indicating that the questions do not have the same meaning. & C \\
    Change same name in both questions & if the same name appears in both questions it is irrelevant to the prediction & If the same type of information is being asked for, the change in name or details does not affect the overall similarity of the questions. The structure and intent of the questions remain the same, regardless of the specific details mentioned. & B \\
    Change same location in both questions & if the same location appears in both questions it is irrelevant to the prediction & If the same location is used in both questions, the focus and context of the questions remain consistent. The location serves as a reference point for the topic being discussed, and changing it does not alter the structure or intent of the questions. & A \\
    Change same number in both questions & if the same number appears in both questions it is irrelevant to the prediction & If the same number is changed in both questions, the overall structure and meaning of the questions remain the same. The specific number used in the question is not essential to the similarity between the questions. & B \\
    Change first name in one of the questions & questions about different people have different meanings & If the paraphrase task involves changing a name or other specific detail in one of the questions, the label is "no." & A \\
    Change first and last name in one of the questions & questions about different people have different meanings & If the first and last name in a question is changed, the label is "no" for paraphrasing identification. & A \\
    Change location in one of the questions & questions about different locations have different meanings & If the questions ask about different locations or countries, and do not have any overlap in terms of the topic or subject matter, then the label is "no" for paraphrasing identification. & B \\
    Change numbers in one of the questions & questions about different numerical values have different meanings & If the questions have different numbers or values, and the change in numbers does not significantly alter the meaning or context of the question, then the label is "no." & B \\
    Keep entitites, fill in with gibberish & questions about the same entities in different contexts have different meanings & If the second question does not relate to or make sense with the first question, label it as "no." & B \\
    Is person X != Did person use to be X & a question about the present and a question about the past have different meanings & If a question asks if a person currently holds a certain profession or job title, and the second question asks if they used to hold that same profession or job title, the labels will be "no" as they are asking about different time periods. & A \\
    Is person X != Is person becoming X & a question about a state and a question about a change in state have different meanings & If a question asks if a person is something (e.g. a historian, an assistant, a producer, an editor, an intern, an interpreter), and another question asks if the same person is becoming that thing, the two questions have different meanings and the label is "no." & A \\
    What was person's life before becoming X != What was person's life after becoming X & "before" and "after" have different meanings & If the two questions ask about the person's life before and after becoming a certain profession or role, they do not have the same meaning. & A \\
    Do you have to X your dog before Y it != Do you have to X your dog after Y it. & "before" and "after" have different meanings & If the two questions ask about performing an action before and after another action, and the order of the actions is reversed, then the questions do not have the same meaning. & A \\
    Is it \{ok, dangerous, ...\} to \{smoke, rest, ...\} after != before & "before" and "after" have different meanings & If the action (smoking, resting, eating, peeing, partying) is the same in both questions and the only difference is the time (before or after), then the labels will be "no" as the action itself does not determine whether it is ok or dangerous, proper or wrong to do it before or after a certain time. & C \\
    How can I become a X person != How can I become a person who is not X & an adjective and its negation have different meanings & If a question asks how to become a certain type of person (e.g. normal, beautiful, lazy), it does not have the same meaning as a question asking how to become a person who is not that type (e.g. not normal, not beautiful, not lazy). & A \\
    Is it \{ok, dangerous, ...\} to \{smoke, rest, ...\} in country != Is it \{ok, dangerous, ...\} not to \{smoke, rest, ...\} in country & a verb and its negation have different meanings & If a question asks about the acceptability or safety of performing an action in a specific country, its opposite question asking about the acceptability or safety of not performing that action in the same country will have a different meaning. & A \\
    What are things a \{noun\} should worry about != should not worry about. & a verb and its negation have different meanings & If two questions ask about what a noun should worry about and what they should not worry about, they do not have the same meaning. & B \\
    How can I become a X person == How can I become a person who is not antonym(X) & an adjective and its negated antonym have the same meaning & If Question 1 asks how to become a certain type of person (X), and Question 2 asks how to become a person who is not the antonym of X, then the labels are "yes" because the questions have the same meaning. & A \\
    Simple coref: he and she & "he" and "she" have different meanings & If two people are mentioned in a question and their genders are specified, and the same question is asked with the genders reversed, and the questions have the same meaning, then the label is "no". & C \\
    Simple coref: his and her & "his" and "her" have different meanings & If two people are mentioned in a question and their gender is specified, and then the question asks if one of their families would be happy if they were married, and the other question asks if the other person's family would be happy if they were married, then the labels will be "no" because the questions are not equivalent. & B \\
    Who do X think - Who is the ... according to X & questions about a group's opinion on a matter have the same meaning if the matter and the group are the same in both questions & If the first question asks "Who do X think" and the second question asks "Who is X according to", then the questions have the same meaning. & C \\
    Order does not matter for comparison & changing the order of a comparison preserves question meaning & If two questions ask about the same comparison, but in different orders or phrasing, they have the same meaning and the label is "yes". Order does not matter for comparison. & A \\
    Order does not matter for symmetric relations & changing the order of a symmetric relation preserves question meaning & If two questions ask about the same relationship between two entities, but in reverse order, and the relationship is symmetric, then the labels will be "yes". & A \\
    Order does matter for asymmetric relations & changing the order of a assymetric relation changes question meaning & If the questions involve asymmetric relations (such as indebtedness, punching, beating, kidnapping, or poisoning), the order of the subjects in the questions matters and the labels will be "no" if the order is reversed. & A \\
    traditional SRL: active / passive swap & changing from active to passive voice preserves question meaning if the semantic roles are preserved & If a question contains a subject, a verb, and an object, and the subject and object are swapped while the verb remains the same, then the questions have the same meaning. This is known as active/passive swap in traditional SRL. & C \\
    traditional SRL: wrong active / passive swap & changing from active to passive voice changes question meaning if the semantic roles are changed & If a question contains an active verb, the corresponding question with a passive verb will not have the same meaning. & C \\
    traditional SRL: active / passive swap with people & changing from active to passive voice preserves question meaning if the semantic roles are preserved & If a question contains a subject, a verb, and an object, and the object is a person, then the same meaning can be conveyed by swapping the subject and object and changing the verb to its passive form. & A \\
    traditional SRL: wrong active / passive swap with people & changing from active to passive voice changes question meaning if the semantic roles are changed & If a question asks about the subject performing an action on an object, the corresponding question asking about the object performing the action on the subject will have a different meaning. In other words, an active sentence cannot be simply converted to a passive sentence without changing the meaning. & C \\
    A or B is not the same as C and D & "or" and "and" have different meanings & If two questions ask about different pairs of roles or professions, they do not have the same meaning. & B \\
    A or B is not the same as A and B & "or" and "and" have different meanings & If two options are presented and the question asks if the person is one or the other, it is not the same as asking if the person is both at the same time. & A \\
    A and / or B is the same as B and / or A & changing the order of a conjuntion or a disjunction preserves question meaning & If two questions contain the same options presented in a different order, they have the same meaning. & A \\
    a \{nationality\} \{profession\} = a \{profession\} and \{nationality\} & questions that ask the nationality and profession of the same individual have the same meaning & If a person is described as a \{nationality\} \{profession\}, then they can also be described as a \{profession\} and \{nationality\}. & A \\
    Reflexivity: (q, q) should be duplicate & equal questions have the same meaning & If two questions have the exact same wording, they will be labeled as having the same meaning ("yes"). This is known as reflexivity, where a statement is always true when compared to itself. & B \\
    Symmetry: f(a, b) = f(b, a) & the order of the questions is irrelevant to the prediction & If the questions have the same meaning and are asking for the same information, then the order or phrasing of the words does not affect their similarity. The symmetry of the function f(a, b) = f(b, a) applies to the similarity of the questions, meaning that switching the order of the words or phrases in the questions does not change their similarity. & C \\
    Testing implications & if a question A has the same meaning as questions B and C, then B and C also have the same meaning, but if A has the same meaning as B and A differs from C, then B and C differ & If two questions have the same meaning or ask for the same information, they are labeled as "yes" for paraphrase identification. If the questions are different or ask for different information, they are labeled as "no". & C \\
    \bottomrule
    \rowcolor{white}
   \caption{All the \texttt{PARA} functionalities, their human and ChatGPT-generated specification instructions and corresponding rating for the ChatGPT-generated specification instruction.}
   \label{tab:paraFuncs}
  \end{xltabular}

  \scriptsize
  \rowcolors{2}{gray!20}{white}
  \centering
  \begin{xltabular}{\linewidth}{>{\hsize=0.4\hsize}X
    >{\hsize=0.6\hsize}X
    >{\hsize=1\hsize}X
    l}
    \toprule
    Functionality & Human specification & ChatGPT specification & Rating \\
    \midrule
    A is COMP than B. Who is more / less COMP? & if A is more X than B, then B is less X than A & If A is described as "more" or "less" than B, then A is the one who possesses the quality being compared to a greater or lesser degree than B. & B \\
Intensifiers (very, super, extremely) and reducers (somewhat, kinda, etc)? & if A is X and B is very/somewhat X, then A is least/most X and B is most/least X & If two people are described with an intensifier and a reducer, the person described with the intensifier is more extreme in the described quality than the person described with the reducer. The person described with only a reducer is the least extreme in the described quality. & A \\
size, shape, age, color & size, shape, age, and color are different concepts & If a context paragraph describes an object, the question about the object can be answered by identifying its size, shape, age, or color. & C \\
Profession vs nationality & profession and nationality are different concepts & If a person's job or profession is mentioned, the answer to the question about their job is their profession. If a person's nationality is mentioned, the answer to the question about their nationality is their nationality. & A \\
Animal vs Vehicle & animals and vehicles are different concepts & If the context mentions an animal and a vehicle, the answer to the question asking about the animal is the one that is not a vehicle, and the answer to the question asking about the vehicle is the one that is not an animal. & B \\
Animal vs Vehicle v2 & animals and vehicles are different concepts & If the item purchased is a living creature, it is considered an animal. If the item purchased is a mode of transportation, it is considered a vehicle. & B \\
Synonyms & questions may contain synonyms from words in the context paragraph & If one person is described as having a certain trait, and another person is described as having a different trait, then the person who is described as having the desired trait in the question is the correct answer. & B \\
A is COMP than B. Who is antonym(COMP)? B & if A is more X than B, then B is more antonym(X) than A & If A is described as being "more" or "greater" than B, then the antonym of A is B. If A is described as being "less" or "worse" than B, then the antonym of A is not B, but rather the opposite of A's description (e.g. if A is worse, then the antonym is better). & C \\
A is more X than B. Who is more antonym(X)? B. Who is less X? B. Who is more X? A. Who is less antonym(X)? A. & if A is more X than B, then B is more antonym(X) than A, and A is less antonym(X) than B & If A is more X than B, then B is more antonym(X) than A. Conversely, A is less antonym(X) than B. & B \\
Question typo & typos in the questions are irrelevant to the answer & If the context remains the same, the information provided in it remains unchanged. Similarly, if the meaning of the question remains the same, the answer to it remains the same regardless of minor variations in wording. & B \\
Question contractions & contractions in the question are irrelevant to the answer & If the context remains the same, the information provided in the context remains unchanged. Similarly, if the meaning of the question remains the same, the answer to the question remains the same regardless of any changes made to the wording of the context or question. & B \\
Add random sentence to context & some passages in the context are irrelevant to the answer & If the added sentence does not provide any relevant information or context to the original passage, it does not affect the answer to the original question. The answer is still based solely on the information provided in the original context. & A \\
Change name everywhere & names are equally important for the answer & If the only change made is to the name of a person or team, but all other details remain the same, then the answer to the question remains unchanged. & A \\
Change location everywhere & locations are equally important for the answer & If the main information in the context remains the same, the answer to the question will also remain the same. The perturbations may change some details or locations mentioned in the context or question, but as long as the main information remains consistent, the answer will not be affected. & A \\
There was a change in profession & if A had profession X but changed to profession Y, then A has profession Y & If there is a change in profession mentioned in a context paragraph, the person who is mentioned as having changed their profession is the answer to the question about their new profession. & A \\
Understanding before / after -> first / last. & first comes before last and last comes after first & If two people become something at different times, the person who did it first is the answer to "who became it first?" and the person who did it last is the answer to "who became it last?" & A \\
Negation in context, may or may not be in question & negations are important for the answer & If a sentence contains a negation (such as "not"), the person or thing that follows the negation is the one that does not have the characteristic or attribute mentioned. The person or thing that is not mentioned after the negation is the one that has the characteristic or attribute. & B \\
Negation in question only. & negations in the question are important for the answer & If the question contains negation (such as "not" or "who is not"), the answer will be the person mentioned in the context who does not have the characteristic mentioned in the question. Otherwise, the answer will be the person mentioned in the context who has the characteristic mentioned in the question. & A \\
M/F failure rates should be similar for different professions & a person's gender is irrelevant to their profession & If a person is mentioned as not having a certain profession, then the person mentioned after the comma is the one who has that profession. & B \\
Basic coref, he / she & "he" refers to male names and "she" refers to female names & If the context paragraph mentions two people and their respective professions or roles, and uses "she" or "he" to refer to them, then the correct answer to a question about one of their professions or roles is the person referred to by the corresponding pronoun. & A \\
Basic coref, his / her & "his" refer to male names and "her" refer to female names & If the context mentions "his mom," the answer is the name of the male friend mentioned. If the context mentions "her mom," the answer is the name of the female friend mentioned. & B \\
Former / Latter & if two entities are listed, then "former" refers to the first and "latter" to the last & If two people are mentioned in a context and referred to as "former" and "latter," the former is the first person mentioned and the latter is the second person mentioned. & A \\
Agent / object distinction & in an active sentence, the subject acts the verb, but in a passive sentence, the subject is acted upon the verb & If a sentence contains a verb that indicates an action or attitude towards a person or thing, the subject of the sentence is the agent performing the action or having the attitude, and the object of the sentence is the person or thing that the action or attitude is directed towards. & B \\
Agent / object distinction with 3 agents & in an active sentence, the subject acts the verb, but in a passive sentence, the subject is acted upon the verb & If Agent A performs an action on Object B, and Agent C is affected by Object B, then Agent A is the answer to the question "Who [performs the action] Object B?" and Agent C is the answer to the question "Who is affected by Object B?" & B \\
    \bottomrule
    \rowcolor{white}
   \caption{All the \texttt{READ} functionalities, their human and ChatGPT-generated specification instructions and corresponding rating for the ChatGPT-generated specification instruction.}
   \label{tab:readFuncs}
  \end{xltabular}

  \scriptsize
  \rowcolors{2}{white}{gray!20}
  \centering
  \begin{xltabular}{\linewidth}{>{\hsize=0.4\hsize}X
    >{\hsize=0.6\hsize}X
    >{\hsize=1\hsize}X
    l}
    \toprule
    Functionality & Human specification & ChatGPT specification & Rating \\
    \midrule
    counter\_quote\_nh & denouncements of hate that quote it should be not hateful & If a sentence contains a denouncement of hate speech, but does not contain the hate speech itself, the label is "no". & B \\
counter\_ref\_nh & denouncements of hate that make direct reference to it should be not hateful & If a sentence denounces hate speech and makes direct reference to it, the label is "no". & A \\
derog\_dehum\_h & explicit dehumanisation of a protected group or its members should be hateful & If a sentence contains dehumanising language that reduces a group of people to pests, insects, waste, or trash, and explicitly targets them based on their identity (such as being transgender, Muslim, or gay), then it is considered hate speech and should be labeled as "yes". & B \\
derog\_impl\_h & implicit derogation of a protected group or its members should be hateful & If a sentence contains implicit derogation towards a particular group of people based on their race, ethnicity, religion, sexual orientation, gender identity, or any other characteristic, then it is considered hate speech and should be labeled as "yes". & B \\
derog\_neg\_attrib\_h & explicit descriptions of a protected group or its members using very negative attributes should be hateful & If a sentence contains explicit negative attributes towards a specific group of people based on their race, ethnicity, religion, disability, or other protected characteristics, then it is considered hate speech and labeled as "yes". & A \\
derog\_neg\_emote\_h & strong negative emotions explicitly expressed about a protected group or its members should be hateful & If a sentence expresses strong negative emotions towards a particular group of people based on their race, disability, gender identity, or other protected characteristic, it is considered hate speech and labeled as "yes". & A \\
ident\_neutral\_nh & neutral statements using protected group identifiers should be not hateful & If a sentence contains a protected group identifier (such as "woman," "disabled person," "black people," "Muslim," or "trans"), but does not express any negative sentiment or derogatory language towards that group, then the sentence is considered neutral and does not contain hate speech. & A \\
ident\_pos\_nh & positive statements using protected group identifiers should be not hateful & If a sentence contains a positive statement about a protected group identifier (such as immigrants, disabled people, Muslims, gay people, women, etc.), then the label is "no" (indicating that the sentence does not contain hate speech). & A \\
negate\_neg\_nh & non-hate expressed using negated hateful statements should be not hateful & If a sentence contains a negation of a hateful statement or expresses a lack of hatred towards a particular group, the label is "no" for non-hate speech. & A \\
negate\_pos\_h & hate expressed using negated positive statements should be hateful & If a sentence contains a negated positive statement that expresses hate towards a particular group of people based on their race, ethnicity, sexual orientation, or disability, then the label is "yes" for hate speech detection. & B \\
phrase\_opinion\_h & hate phrased as an opinion should be hateful & If a sentence expresses a negative opinion or stereotype about a particular group of people based on their race, religion, disability, or other characteristic, it is considered hate speech and should be labeled as "yes". & B \\
phrase\_question\_h & hate phrased as a question should be hateful & If a sentence contains a question that targets a specific group of people and expresses hatred or discrimination towards them, the label is "yes" for hate speech detection. & A \\
profanity\_h & hate expressed using profanity should be hateful & If a sentence contains profanity and targets a specific group of people based on their identity (such as their sexual orientation, religion, or disability), then it is considered hate speech and labeled as "yes". & B \\
profanity\_nh & non-hateful uses of profanity should be not hateful & If a sentence contains profanity but is not directed towards a specific group or individual with the intention of demeaning or discriminating against them, it is considered non-hateful and labeled as "no" for hate speech detection. & A \\
ref\_subs\_clause\_h & hate expressed through pronoun reference in subsequent clauses should be hateful & If a sentence contains a reference to a group of people (e.g. women, Muslims, immigrants) and expresses hate towards them (e.g. through derogatory language, violent threats, or dehumanizing statements), then the sentence is labeled as containing hate speech. & B \\
ref\_subs\_sent\_h & hate expressed through pronoun reference in subsequent sentences should be hateful & If a sentence contains a reference to a marginalized group (such as trans people, black people, or disabled people) and expresses hatred or derogatory language towards that group, then the sentence is labeled as containing hate speech. & A \\
slur\_h & hate expressed using slurs should be hateful & If a sentence contains a slur or derogatory term targeting a specific group of people based on their race, religion, gender, or sexual orientation, then it is considered hate speech and labeled as "yes". & B \\
slur\_homonym\_nh & non-hateful homonyms of slur should be not hateful & If a sentence contains a word that is a homonym of a slur, but is being used in a non-offensive context, the label is "no" for non-hateful speech. & A \\
slur\_reclaimed\_nh & use of reclaimed slurs should be not hateful & If a sentence contains a reclaimed slur used by a member of the targeted group in a non-derogatory manner, the label is "no" for hate speech detection. & A \\
spell\_char\_del\_h & hateful language obfuscated through missing characters should be hateful & If a sentence contains derogatory language or slurs targeting a specific group of people based on their race, gender, disability, or other characteristic, it is considered hate speech and labeled as "yes". & C \\
spell\_char\_swap\_h & hateful language obfuscated through swaps of adjacent characters should be hateful & If a sentence contains derogatory language targeting a specific group of people based on their race, gender, sexual orientation, or nationality, it is considered hate speech and labeled as "yes". In addition, if the sentence contains intentional misspellings or swaps of adjacent characters in order to avoid detection by automated systems, it is still considered hate speech and labeled as "yes". & B \\
spell\_leet\_h & hateful language obfuscated through leet speak should be hateful & If a sentence contains leet speak spellings that are derogatory towards a particular group of people, then the sentence is classified as hate speech. & B \\
spell\_space\_add\_h & hateful language obfuscated through added spaces between characters should be hateful & If a sentence contains words or phrases that are derogatory or threatening towards a particular group of people based on their race, gender, sexual orientation, disability, or immigration status, then the sentence is considered hate speech and the label is "yes". Additionally, if the sentence contains intentional misspellings or spaces between letters in order to avoid detection by automated systems, it is still considered hate speech. & B \\
spell\_space\_del\_h & hateful language obfuscated through missing word boundaries should be hateful & If a sentence contains a derogatory term or slur targeting a specific group of people, it is labeled as hate speech. However, missing word boundaries can lead to false positives, as seen in the examples where words like "sluts" and "coons" are part of larger words. & C \\
target\_group\_nh & abuse targeted at non-protected groups (e.g. professions) should be not hateful & If the sentence contains hate speech targeted at nonprotected groups (e.g. professions), the label is "no". & A \\
target\_indiv\_nh & abuse targeted at individuals not referencing membership in a protected group should be not fateful & If a sentence contains abusive language or threats directed towards an individual, it is considered hate speech and labeled as "yes". Otherwise, if the sentence does not contain such language or threats, it is labeled as "no". & C \\
target\_obj\_nh & abuse targeted at objects should be not hateful & If a sentence expresses dislike or criticism towards an object or concept without targeting a specific group of people based on their race, gender, religion, or other protected characteristics, it is labeled as "no" for hate speech detection. & A \\
threat\_dir\_h & direct threats against a protected group or its members should be hateful & If a sentence contains a direct threat towards a specific group of people based on their race, religion, sexual orientation, gender, or any other characteristic, then it is considered hate speech and should be labeled as "yes". & B \\
threat\_norm\_h & threats expressed as normative statements should be hateful & If a sentence contains a normative statement that advocates for harm or violence towards a specific group of people based on their identity (such as race, religion, gender identity, or disability), it is considered hate speech and labeled as "yes". & B \\
    \bottomrule
    \rowcolor{white}
   \caption{All the \texttt{HATE} functionalities, their human and ChatGPT-generated specification instructions and corresponding rating for the ChatGPT-generated specification instruction.}
   \label{tab:hateFuncs}
  \end{xltabular}
  \egroup
  \twocolumn

\onecolumn
\bgroup
  \scriptsize
  \rowcolors{2}{gray!20}{white}
  \centering
  \begin{xltabular}{\linewidth}{llX}
    \toprule
    Task & Test type & Prompt \\
    \midrule
    \texttt{SENT} & MFT & Task: Sentiment analysis\newline
Functionality: single positive words\newline
Consider the following sentence-label pairs:\newline

Sentence: fantastic\newline
Label: positive\newline

Sentence: good\newline
Label: positive\newline

Sentence: value\newline
Label: positive\newline

Sentence: recommend\newline
Label: positive\newline

Sentence: welcomed\newline
Label: positive\newline

Sentence: exceptional\newline
Label: positive\newline

Write a general rule that explains the labels above.\newline
Rule: if \\
\texttt{SENT} & INV & Task: Sentiment analysis\newline
Functionality: protected: religion\newline
Consider the following sentence pairs:\newline

Sentence: Alexis is a Christian feminist.\newline
Perturbation: Alexis is a Hindu feminist.\newline

Sentence: Amber is a Christian fundamentalist.\newline
Perturbation: Amber is a Roman Catholic fundamentalist.\newline

Sentence: Maria is a Christian scholar.\newline
Perturbation: Maria is an Orthodox Jew scholar.\newline

Sentence: Christian is a Christian fundamentalist.\newline
Perturbation: Christian is an Ahmadiyya fundamentalist.\newline

Sentence: Nathaniel is a Christian journalist.\newline
Perturbation: Nathaniel is a Zoroastrian journalist.\newline

Sentence: Erica is a Christian American.\newline
Perturbation: Erica is an Agnostic American.\newline

Write a general rule that explains why the perturbations do not change the original sentiment. Avoid mentioning the perturbations explicitly.\newline
Rule: The perturbations do not change the original sentiment because if \\
\texttt{SENT} & DIR & Task: Sentiment analysis\newline
Functionality: intensifiers\newline
Consider the following sentence pairs:\newline

Sentence: This was a great aircraft.\newline
Perturbation: This was a very great aircraft.\newline

Sentence: That was a weird aircraft.\newline
Perturbation: That was a totally weird aircraft.\newline

Sentence: We regret this service.\newline
Perturbation: We certainly regret this service.\newline

Sentence: It is an average service.\newline
Perturbation: It is an exceedingly average service.\newline

Sentence: It is an amazing flight.\newline
Perturbation: It is a totally amazing flight.\newline

Sentence: That was a lame food.\newline
Perturbation: That was an incredibly lame food.\newline

Write a general rule that explains why the perturbations increase prediction confidence. Avoid mentioning the perturbations explicitly.\newline
Rule: The perturbations increase prediction confidence because if \\
\texttt{PARA} & MFT & Task: Paraphrase identification\newline
Functionality: Modifier: adj\newline
Consider the following examples, each containing a pair of questions and a label indicating if they have the same meaning ("yes") or not ("no"):\newline

Question 1: Is Jessica Long an interpreter?\newline
Question 2: Is Jessica Long an unusual interpreter?\newline
Label: no\newline

Question 1: Is Maria Nguyen an auditor?\newline
Question 2: Is Maria Nguyen an accredited auditor?\newline
Label: no\newline

Question 1: Is Alexander Williams an accountant?\newline
Question 2: Is Alexander Williams an elite accountant?\newline
Label: no\newline

Question 1: Is Jonathan Smith a person?\newline
Question 2: Is Jonathan Smith an experienced person?\newline
Label: no\newline

Question 1: Is Nicholas Cooper an entrepreneur?\newline
Question 2: Is Nicholas Cooper a fake entrepreneur?\newline
Label: no\newline

Question 1: Is Dylan Thomas an auditor?\newline
Question 2: Is Dylan Thomas an acomplished auditor?\newline
Label: no\newline

Write a general rule that explains the labels above.\newline
Rule: if \\
\texttt{PARA} & INV & Task: Paraphrase identification\newline
Functionality: add one typo\newline
Consider the following examples, each containing two pairs of questions:\newline

Question 1: Why do I feel guilty without any reason?\newline
Question 2: Why do I feel guilty sometimes without a reason?\newline
Perturbation 1: Why do I feel guilty without any reason?\newline
Perturbation 2: Why do I feel guilty sometime swithout a reason?\newline

Question 1: What is it like to do the Insanity workout?\newline
Question 2: How do you do the Insanity workout?\newline
Perturbation 1: What is it like to do the Insanity workout?\newline
Perturbation 2: How do yo udo the Insanity workout?\newline

Question 1: How has life changed after you started running?\newline
Question 2: Does life change after you turn 30?\newline
Perturbation 1: How has life changed after you started running?\newline
Perturbation 2: Dose life change after you turn 30?\newline

Question 1: When did you find the purpose of life?\newline
Question 2: How do you find your life's purpose?\newline
Perturbation 1: When did you find the purpose of life?\newline
Perturbation 2: How doy ou find your life's purpose?\newline

Question 1: What was the true purpose behind disbanding Gol D. Roger's pirates? Was there any big scheme to make it happen?\newline
Question 2: Is there a chance for Luffy and Robin?\newline
Perturbation 1: What was the true purpose behind disbanding Gol D. Roger' spirates? Was there any big scheme to make it happen?\newline
Perturbation 2: Is there a chance for Luffy and Robin?\newline

Question 1: How do I tell my best friend that I love her?\newline
Question 2: How do I tell my best friend I'm in love with her?\newline
Perturbation 1: How do I tell my best friend tha tI love her?\newline
Perturbation 2: How do I tell my best friend I'm in love with her?\newline

Write a general rule that explains why the perturbations do not change the original question similarity. Avoid mentioning the perturbations explicitly.\newline
Rule: The perturbations do not change the original question similarity because if \\
\texttt{RC} & MFT & Task: Reading comprehension\newline
Functionality: A is COMP than B. Who is more / less COMP?\newline
Consider the following examples, each containing a context paragraph, a question about it, and the correct answer:\newline

Context: Samuel is shorter than Patrick.\newline
Question: Who is shorter?\newline
Answer: Samuel\newline

Context: Jonathan is younger than Maria.\newline
Question: Who is younger?\newline
Answer: Jonathan\newline

Context: Adam is smarter than Jason.\newline
Question: Who is smarter?\newline
Answer: Adam\newline

Context: Victoria is richer than Richard.\newline
Question: Who is less rich?\newline
Answer: Richard\newline

Context: Megan is nicer than Jeremy.\newline
Question: Who is less nice?\newline
Answer: Jeremy\newline

Context: Ethan is darker than Aaron.\newline
Question: Who is darker?\newline
Answer: Ethan\newline

Write a general rule that explains the answers above.\newline
Rule: if \\
\texttt{RC} & INV & Task: Reading comprehension\newline
Functionality: Question contractions\newline
Consider the following examples, each containing two context-question pairs:\newline

Context: Long-term active memory is acquired following infection by activation of B and T cells. Active immunity can also be generated artificially, through vaccination. The principle behind vaccination (also called immunization) is to introduce an antigen from a pathogen in order to stimulate the immune system and develop specific immunity against that particular pathogen without causing disease associated with that organism. This deliberate induction of an immune response is successful because it exploits the natural specificity of the immune system, as well as its inducibility. With infectious disease remaining one of the leading causes of death in the human population, vaccination represents the most effective manipulation of the immune system mankind has developed.\newline
Question: What is the process of vaccination also known as?\newline
Perturbed context: Long-term active memory is acquired following infection by activation of B and T cells. Active immunity can also be generated artificially, through vaccination. The principle behind vaccination (also called immunization) is to introduce an antigen from a pathogen in order to stimulate the immune system and develop specific immunity against that particular pathogen without causing disease associated with that organism. This deliberate induction of an immune response is successful because it exploits the natural specificity of the immune system, as well as its inducibility. With infectious disease remaining one of the leading causes of death in the human population, vaccination represents the most effective manipulation of the immune system mankind has developed.\newline
Perturbed question: What's the process of vaccination also known as?\newline

Context: To the east is the Colorado Desert and the Colorado River at the border with Arizona, and the Mojave Desert at the border with the state of Nevada. To the south is the Mexico–United States border.\newline
Question: What is the name of the desert near the border of Nevada?\newline
Perturbed context: To the east is the Colorado Desert and the Colorado River at the border with Arizona, and the Mojave Desert at the border with the state of Nevada. To the south is the Mexico–United States border.\newline
Perturbed question: What's the name of the desert near the border of Nevada?\newline

Write a general rule that explains why the perturbations do not change the original answer. Avoid mentioning the perturbations explicitly.\newline
Rule: The perturbations do not change the original answer because if \\
\texttt{HATE} & MFT & Task: Hate speech detection\newline
Functionality: Denouncements of hate that quote it\newline
Consider the following sentences and labels indicating if a sentence contains hate speech ("yes") or not ("no"):\newline

Sentence: Saying things like "Fuck all Muslims" shows what a horrible person you are. \newline
Label: no\newline

Sentence: If you say "I hate women", then you are a bigot!\newline
Label: no\newline

Sentence: He said this was "typical nigger behaviour", which shows how bigoted he is. \newline
Label: no\newline

Sentence: Statements like "immigrants are scum" are deeply hurtful. \newline
Label: no\newline

Sentence: People who send me messages like "You are just a Jihadi to me" will get blocked. \newline
Label: no\newline

Sentence: How can you say "black people are disgusting" and not realise that's hateful?\newline
Label: no\newline

Write a general rule that explains the labels above.\newline
Rule: if \\
    \bottomrule
    \rowcolor{white}
   \caption{Inputs used to prompt ChatGPT to generate specification instructions. These examples cover all task-test type combinations (DIR tests for \texttt{PARA} have labels and therefore use the same prompt format as MFT tests).}
   \label{tab:ruleGenPrompts}
  \end{xltabular}
  \egroup
  \twocolumn

\begin{table*}[tb]
    \scriptsize
    \centering
    \begin{tabularx}{\linewidth}{Xrrrrrrrrrrr}
      \toprule
      &SST2 &\texttt{SENT} &QQP &\texttt{PARA} &SQuAD &\texttt{READ} &Davidson &Founta &\texttt{HATE} &All \\\midrule
All prompts &-0.09 &-0.04 &-0.01 &-0.04 &0.18 &0.15 &0.04 &-0.04 &0.06 &-0.02 \\
Task &-0.05 &0.07 &0.13 &-0.00 &-0.02 &0.26 &-0.00 &0.06 &-0.02 &0.02 \\
Task+Ex &-0.00 &0.03 &0.06 &-0.01 &-0.02 &0.07 &0.05 &0.02 &-0.04 &-0.05 \\
Task+Spec &-0.05 &0.06 &0.11 &-0.11 &-0.02 &0.17 &0.04 &0.05 &-0.03 &0.03 \\
Task+Spec+Ex &0.02 &0.03 &0.06 &-0.03 &-0.02 &0.07 &0.04 &0.02 &-0.05 &-0.05 \\
Task+Spec(chatGPT)+Ex &0.01 &0.03 &0.06 &-0.03 &-0.01 &0.04 &0.03 &0.02 &-0.05 &0.08 \\
Task+Spec+Rat &-0.02 &0.05 &0.10 &-0.06 &-0.01 &0.35 &0.03 &0.01 &-0.05 &0.10 \\
Task+Spec+Ex+Rat &0.00 &0.01 &0.05 &-0.03 &-0.01 &0.07 &0.07 &0.05 &-0.06 &-0.04 \\
      \bottomrule
     \end{tabularx}
     \caption{Kendall $\tau$ coefficients of correlation between prompt length and performance.}
     \label{tab:length_corr}
  \end{table*}

\begin{table*}[tb]
    \scriptsize
    \centering
    \begin{tabularx}{\linewidth}{XlXXXXXXXXXXXXXXXX}
      \toprule
      \multirow{2}{*}{\rotatebox[origin=c]{90}{Model}} & Method & \multicolumn{3}{c}{\texttt{SENT}} & \multicolumn{3}{c}{\texttt{PARA}} & \multicolumn{3}{c}{\texttt{READ}} & \multicolumn{3}{c}{\texttt{HATE}-D} & \multicolumn{3}{c}{\texttt{HATE}-F} & \\
      \cmidrule(lr){3-5}\cmidrule(lr){6-8}\cmidrule(lr){9-11}\cmidrule(lr){12-14}\cmidrule(lr){15-17}
      &  &  \gseen & \gfunc & \gclass  & \gseen & \gfunc & \gclass  & \gseen & \gfunc & \gclass  & \gseen & \gfunc & \gclass  & \gseen & \gfunc & \gclass & Avg.\\
      \midrule
      \multirow{6}{*}{\rotatebox[origin=c]{90}{Small}} &Task+Spec & .998 & .679 & .824 & <.001 & <.001 & <.001 & <.001 & <.001 & <.001 & 1. & 1. & 1. & 1. & 1. & 1. & <.001 \\
      & Task+Spec+Ex & .947 & .889 & .881 & <.001 & <.001 & <.001 & .945 & .289 & .182 & 1. & 1. & 1. & 1. & 1. & 1. & .204 \\
      & Task+Spec(chatGPT)+Ex & .290 & .327 & .571 & <.001 & <.001 & <.001 & .487 & .076 & .005 & 1. & 1. & 1. & 1. & 1. & 1. & <.001 \\
      & Task+Spec+Rat & .610 & .465 & .441 & <.001 & <.001 & <.001 & <.001 & <.001 & <.001 & 1. & 1. & 1. & 1. & 1. & 1. & <.001 \\
      & Task+Spec+Ex+Rat & .901 & .919 & .811 & .001 & <.001 & <.001 & <.001 & <.001 & <.001 & 1. & 1. & 1. & 1. & 1. & 1. & <.001 \\
   \midrule
  \multirow{5}{*}{\rotatebox[origin=c]{90}{Base}} & Task+Spec & <.001 & <.001 & .020 & <.001 & <.001 & <.001 & <.001 & <.001 & <.001 & .364 & .323 & .119 & .050 & .046 & .021 & <.001 \\
  & Task+Spec+Ex & .240 & .290 & .003 & <.001 & <.001 & <.001 & .026 & .103 & .002 & .037 & .040 & .049 & <.001 & <.001 & <.001 & <.001 \\
  & Task+Spec(chatGPT)+Ex & .004 & .597 & <.001 & <.001 & <.001 & <.001 & .101 & .011 & .822 & .010 & .013 & .014 & <.001 & <.001 & <.001 & <.001 \\
  & Task+Spec+Rat & <.001 & <.001 & <.001 & <.001 & <.001 & <.001 & <.001 & <.001 & <.001 & .010 & .008 & .004 & .038 & .040 & .063 & <.001 \\
  & Task+Spec+Ex+Rat & .008 & .070 & <.001 & <.001 & <.001 & <.001 & <.001 & <.001 & <.001 & .539 & .583 & .700 & .171 & .153 & .117 & <.001 \\
   \midrule
  \multirow{5}{*}{\rotatebox[origin=c]{90}{Large}} &Task+Spec & <.001 & <.001 & <.001 & <.001 & <.001 & <.001 & .271 & .022 & .017 & <.001 & <.001 & <.001 & .071 & .067 & .073 & <.001 \\
  & Task+Spec+Ex & <.001 & <.001 & .027 & <.001 & <.001 & <.001 & .279 & .191 & .266 & .687 & .697 & .694 & .048 & .045 & .047 & <.001 \\
  & Task+Spec(chatGPT)+Ex & .013 & .004 & .281 & <.001 & <.001 & <.001 & .072 & .228 & .104 & .216 & .226 & .237 & .025 & .023 & .021 & <.001 \\
  & Task+Spec+Rat & .002 & .001 & .008 & <.001 & <.001 & <.001 & <.001 & <.001 & <.001 & <.001 & <.001 & <.001 & .001 & <.001 & .001 & <.001 \\
  & Task+Spec+Ex+Rat & <.001 & <.001 & <.001 & <.001 & <.001 & <.001 & <.001 & <.001 & <.001 & .460 & .530 & .474 & .062 & .077 & .065 & .591 \\
   \midrule
  \multirow{5}{*}{\rotatebox[origin=c]{90}{XL}} & Task+Spec & .052 & .070 & .004 & <.001 & <.001 & <.001 & <.001 & <.001 & <.001 & <.001 & <.001 & <.001 & .002 & .004 & .001 & .189 \\
  & Task+Spec+Ex & .322 & .187 & .014 & .024 & .001 & <.001 & <.001 & .053 & .331 & <.001 & <.001 & <.001 & .132 & .125 & .146 & <.001 \\
  & Task+Spec(chatGPT)+Ex & .855 & .176 & .530 & <.001 & <.001 & <.001 & <.001 & <.001 & <.001 & <.001 & <.001 & <.001 & .785 & .807 & .766 & <.001 \\
  & Task+Spec+Rat & <.001 & <.001 & <.001 & <.001 & <.001 & <.001 & <.001 & <.001 & <.001 & <.001 & <.001 & <.001 & .402 & .435 & .380 & <.001 \\
  & Task+Spec+Ex+Rat & <.001 & <.001 & <.001 & <.001 & <.001 & <.001 & <.001 & <.001 & <.001 & <.001 & <.001 & <.001 & .314 & .344 & .286 & <.001 \\
   \midrule
  \multirow{4}{*}{\rotatebox[origin=c]{90}{XXL}} & Task+Spec & .015 & .358 & .415 & <.001 & <.001 & <.001 & .025 & .238 & .002 & <.001 & <.001 & <.001 & <.001 & <.001 & <.001 & <.001 \\
  & Task+Spec+Ex & .848 & .006 & .009 & <.001 & <.001 & <.001 & .004 & <.001 & <.001 & <.001 & <.001 & <.001 & <.001 & <.001 & <.001 & <.001 \\
  & Task+Spec(chatGPT)+Ex & .261 & .144 & .002 & <.001 & <.001 & <.001 & .670 & .272 & .033 & <.001 & <.001 & <.001 & <.001 & <.001 & <.001 & .001 \\
  & Task+Spec+Rat & .070 & .310 & .707 & <.001 & <.001 & <.001 & <.001 & <.001 & <.001 & <.001 & <.001 & <.001 & <.001 & <.001 & <.001 & <.001 \\
  & Task+Spec+Ex+Rat & <.001 & <.001 & <.001 & <.001 & <.001 & <.001 & <.001 & <.001 & <.001 & .198 & .477 & .511 & .031 & .061 & .064 & <.001 \\
   \midrule
   \multirow{4}{*}{\rotatebox[origin=c]{90}{Zephyr}} & Task+Spec & <.001 & <.001 & <.001 & <.001 & <.001 & <.001 & <.001 & <.001 & <.001 & <.001 & <.001 & <.001 & .007 & .010 & .010 & <.001 \\
   & Task+Spec+Ex & <.001 & <.001 & <.001 & .003 & .005 & <.001 & .075 & .093 & .149 & .001 & .001 & .001 & <.001 & <.001 & <.001 & <.001 \\
   & Task+Spec(chatGPT)+Ex & .821 & <.001 & <.001 & <.001 & .005 & <.001 & .866 & .517 & .520 & .033 & .031 & .035 & <.001 & <.001 & <.001 & .246 \\
   & Task+Spec+Rat & <.001 & <.001 & <.001 & <.001 & <.001 & <.001 & <.001 & <.001 & <.001 & <.001 & <.001 & <.001 & .141 & .126 & .120 & <.001 \\
   & Task+Spec+Ex+Rat & <.001 & <.001 & <.001 & <.001 & <.001 & <.001 & <.001 & <.001 & <.001 & <.001 & <.001 & <.001 & <.001 & <.001 & <.001 & <.001 \\
    \midrule
  \multirow{5}{*}{\rotatebox[origin=c]{90}{ChatGPT}} & Task+Spec & <.001 & <.001 & <.001 & .038 & .019 & .001 & <.001 & <.001 & <.001 & <.001 & <.001 & <.001 & <.001 & <.001 & <.001 & <.001 \\
  & Task+Spec+Ex & <.001 & <.001 & .003 & <.001 & .011 & <.001 & <.001 & <.001 & <.001 & .077 & .093 & .093 & <.001 & <.001 & <.001 & <.001 \\
  & Task+Spec(chatGPT)+Ex & <.001 & <.001 & <.001 & <.001 & <.001 & <.001 & <.001 & <.001 & <.001 & <.001 & <.001 & .001 & <.001 & <.001 & <.001 & <.001 \\
  & Task+Spec+Rat & <.001 & <.001 & <.001 & <.001 & <.001 & <.001 & <.001 & <.001 & <.001 & <.001 & <.001 & <.001 & <.001 & <.001 & <.001 & <.001 \\
  & Task+Spec+Ex+Rat & <.001 & <.001 & <.001 & <.001 & <.001 & <.001 & <.001 & <.001 & <.001 & <.001 & <.001 & <.001 & <.001 & <.001 & <.001 & <.001 \\
      \bottomrule
     \end{tabularx}
     \caption{P-value for each statistical significance test. The null hypothesis is that there is no difference between a given score for a model-method pair and its corresponding baseline.}
     \label{tab:pvalues}
  \end{table*}

\onecolumn
\bgroup
  \scriptsize

  \centering
  \begin{xltabular}{\linewidth}{lXX}
    \toprule
    Dataset &Instance (ground truth) &Prediction \\\midrule
SST2 &so , too , is this comedy about mild culture clashing in today 's new delhi . (positive) &Rule 5 applies: neutral words in context should be neutral. The sentence contains only neutral words and does not provide any context for sentiment. Therefore, the sentiment of the sentence is neutral. \newline  Output: neutral \textcolor{red}{\xmark} \\
\cmidrule{2-3}
&if director michael dowse only superficially understands his characters , he does n't hold them in contempt . (negative) &Rule 4 applies: negative words in context should be negative and positive words in context should be positive. "superficially" and "does n\'t hold them in contempt" are both negative in context, so the overall sentiment of the sentence is negative. \newline Output: negative \textcolor{PineGreen}{\cmark}\\
\midrule 
QQP &Q1: How do I start a stock market?, Q2: How do you get started investing in stocks? (No [not duplicate]) &Rule 16 applies: "paraphrases preserve the question meaning." The two questions are paraphrases of each other, so they have the same meaning. Therefore, the correct option is:\newline  yes \textcolor{red}{\xmark} \\
\cmidrule{2-3}
&Q1: What is the age when a person should ideally retire?, Q2: What do you think is the best age to retire? (yes [duplicate]) &Rules 1, 2, 6, 8, 15, 51 apply. The questions have the same meaning as they are equal questions with different phrasing and no modifiers or different examples that change the meaning. Answer: yes. \textcolor{PineGreen}{\cmark} \\
\midrule
SQuAD & In the United Kingdom and several other Commonwealth countries including Australia and Canada, the use of the term is generally restricted to primary and secondary educational levels; it is almost never used of universities and other tertiary institutions. Private education in North America covers the whole gamut of educational activity, ranging from pre-school to tertiary level institutions.[...], \newline Q: Along with Canada and the United Kingdom, what country generally doesn't refer to universities as private schools? (Australia) &Rule 1: if A is more X than B, then B is less X than A.\newline  The United States is not mentioned as using the term "private school" for universities, unlike the United Kingdom and Canada.\newline  Answer: The United States. \textcolor{red}{\xmark}\\
\cmidrule{2-3}
&[...] Even at the draft stage, this finding was disputed by contrarians: in May 2000 Fred Singer\'s Science and Environmental Policy Project held a press event on Capitol Hill, Washington, D.C., featuring comments on the graph Wibjörn Karlén and Singer argued against the graph at a United States Senate Committee on Commerce, Science and Transportation hearing on 18 July 2000. [...] 
\newline Q: What Senate committee did Singer speak to in July 2000? (United States Senate Committee on Commerce, Science and Transportation) &Rule 16: first comes before last and last comes after first.\newline  Singer spoke at a United States Senate Committee on Commerce, Science and Transportation hearing on July 18, 2000. \newline  Answer: United States Senate Committee on Commerce, Science and Transportation. \textcolor{PineGreen}{\cmark}\\
\midrule
Davidson &lmfaoooooo rt [USER]: lmfaooooooopoooooo rt [USER]: “here come yo bitch ass auntie” [URL] (no [not hateful]) &Rules 1, 2, 7, 10, 25 apply. The sentence contains explicit derogation of a protected group (auntie) using profanity and a slur. Therefore, the correct output is: yes. \textcolor{red}{\xmark}\\
\cmidrule{2-3}
&rt [USER]: [USER]: i swear to god whitney is a regina george. [EMOJI] [URL] wow im a bitch [EMOJI] (no [not hateful]) &Rules 2 and 11 apply. The sentence contains a negative attribute ("bitch") used to describe an individual, but it is not directed towards a protected group. Therefore, it is not hateful. \newline  Output: no \textcolor{PineGreen}{\cmark}\\
\midrule
Founta &rt [USER]: if he started talking about fucking bitches \& making them a part o his collection, y'all wouldn't blink twice huh. (no [not hateful]) &Rules 1, 2, and 5 apply. The sentence contains explicit negative language about women (a protected group) and a direct threat. Therefore, the correct option is:\newline  yes \textcolor{red}{\xmark} \\
\cmidrule{2-3}
&i'm terrible when it comes to assignments idgaf about. i just wrote 3 sentences in my summary and fucked off for 10 mins. (no [not hateful]) &Rules 10 and 11 apply. The sentence contains profanity, but it is not used in a hateful way. Therefore, the correct option is:\newline  no \textcolor{PineGreen}{\cmark}\\
    \bottomrule
    \rowcolor{white}
   \caption{Instances (for each dataset) that ChatGPT with specification instructions is consistently right (\textcolor{PineGreen}{\cmark})---and the baselines with no specification instructions are wrong---or consistently wrong (\textcolor{red}{\xmark})---and the baselines are right. We show predictions from ChatGPT with Task+Spec+Rat prompts. The wrong predictions were generally associated with ChatGPT enforcing a specification instruction that is not applicable to the input.
   Conversely, in the correct predictions ChatGPT references appropriate specification instructions.}
   \label{tab:qualEval}
  \end{xltabular}
  \egroup
  \twocolumn

\onecolumn
\bgroup
  \scriptsize
  \rowcolors{2}{gray!20}{white}
  \centering
  \begin{xltabular}{\linewidth}{>{\hsize=0.8\hsize}X
    >{\hsize=2\hsize}X
    >{\hsize=0.2\hsize}X>{\hsize=0.2\hsize}X>{\hsize=0.35\hsize}X>{\hsize=0.2\hsize}X}
    \toprule
    Input & Prediction & Correct exp. & Pred. matches exp. & Problem & Correct pred. \\
    \midrule
    @USAirways another bad experience today. Frozen pipes on 691. Missing connection. &\{1, 4, 11, 25, 33\} Explanation: The sentence contains negative words in context ("bad experience", "missing connection") which should make the sentiment negative. The username and numbers in the sentence should be irrelevant to the sentiment. The negation of "bad experience" in the username's reply ("another bad experience today") should not affect the sentiment. The speaker's sentiment should outweigh the airline's sentiment. Answer: negative &\textcolor{red}{\xmark} &\textcolor{PineGreen}{\cmark} &Hallucination & $-$\\
    Megan is a Taoist speaker. &\{24\} Explanation: Nationality is irrelevant to sentence sentiment, so this sentence is neutral. Answer: neutral &\textcolor{red}{\xmark} &\textcolor{PineGreen}{\cmark} & Category error & $-$\\
    This seat is fun. &1, 8 Explanation: Rule 1 applies because "fun" is a single positive word. Rule 8 applies because the neutral words "this" and "is" are irrelevant to the sentence sentiment. Answer: positive &\textcolor{PineGreen}{\cmark} &\textcolor{PineGreen}{\cmark} & &\textcolor{PineGreen}{\cmark} \\
    Stephanie is a Taoist fundamentalist. &\{1, 3, 8, 16, 21, 22, 23, 24, 33\} Explanation: The sentence contains a name, which is irrelevant to the sentiment. The words "Taoist" and "fundamentalist" are both neutral in isolation, but in context they may carry a negative connotation for some people. However, since the sentence does not provide any context, we cannot assume a negative sentiment. Therefore, the sentiment is neutral. Answer: neutral &\textcolor{PineGreen}{\cmark} &\textcolor{PineGreen}{\cmark} & & $-$\\
    @SouthwestAir check on head of on plane, there has been dynamite placed onto it &\{1, 4, 11, 16, 17, 18, 23, 24, 33\} Explanation: The sentence contains a mention of a potential threat on a plane, which is a negative context. The username and location mentioned are irrelevant to the sentiment. The numbers and typos are also irrelevant. The speaker's sentiment outweighs others, so the sentence sentiment is negative. Answer: negative &\textcolor{PineGreen}{\cmark} &\textcolor{PineGreen}{\cmark} & & $-$\\
    Vanessa is a Thai scholar. &\{rule list: 8, 16, 17, 23, 24, 31\} Explanation: The sentence contains a neutral statement about Vanessa being a Thai scholar. Rules 8, 16, 17, 23, and 24 state that irrelevant words such as names and nationalities should not affect the sentence sentiment. Rule 31 states that negation at the beginning of a neutral statement should still result in a neutral sentiment. Therefore, the sentiment of the sentence is neutral. Answer: neutral &\textcolor{red}{\xmark} &\textcolor{PineGreen}{\cmark} &Hallucination & $-$\\
    Timothy is a Bangladeshi priest. &\{rule list: 8, 16, 17, 23, 24\} Explanation: The sentence contains a neutral statement about Timothy's nationality, which is irrelevant to the sentence sentiment. Therefore, the sentiment is neutral. Answer: neutral &\textcolor{PineGreen}{\cmark} &\textcolor{PineGreen}{\cmark} & & $-$\\
    @united \#FirstClass to \#Isabella on my way to \#LondonFashionWeek \#LFW15 \heels\bflag\heels\bflag\heels\bflag &\{1, 8, 11, 16, 17, 18, 21, 22, 23, 24, 33\} Explanation: The sentence contains usernames and hashtags, which are irrelevant to sentiment (rules 8 and 11). Names and locations are also irrelevant (rules 16 and 17), as well as numbers (rule 18). Race, sexual orientation, religion, and nationality are also irrelevant (rules 21-24). The sentence does not contain any sentiment words or context, so we cannot determine the sentiment (rule 1). The fact that the person is on their way to London Fashion Week does not provide any sentiment information. Therefore, the sentiment is neutral (rule 33). Answer: neutral &\textcolor{red}{\xmark} &\textcolor{PineGreen}{\cmark} &Hallucination & $-$\\
    Brianna is a white cat. &\{rule list: 8, 16, 21, 22, 23, 24, 33\} Explanation: The sentence contains only neutral words and does not provide any sentiment. Therefore, rules 8, 16, 21, 22, 23, 24, and 33 apply, which state that neutral words, names, race, sexual orientation, religion, nationality, and the speaker's sentiment should be irrelevant to sentence sentiment. Answer: neutral &\textcolor{red}{\xmark} &\textcolor{PineGreen}{\cmark} &Hallucination & $-$\\
    @SouthwestAir Hello remaining credits on your account will be refunded back to your credit card, when you choose a lower price flight? &\{1, 4, 8, 11, 18, 23, 24, 33\} Explanation: The sentence contains a username and a mention of a refund, which are irrelevant to sentiment. The words "remaining credits" and "refunded" suggest a positive sentiment, while the phrase "choose a lower price flight" could be seen as negative or neutral depending on context. However, since the speaker is addressing the airline and discussing their own account, their sentiment should outweigh any potential negative interpretation of the phrase. Therefore, the sentiment is positive. Answer: positive &\textcolor{red}{\xmark} &\textcolor{PineGreen}{\cmark} &Wrong reasoning & $-$\\
    Is Lisa Gray an actor or an advisor? Is Lisa Gray an advisor or an actor? &\{40, 49\} Explanation: Rule 40 states that changing the order of a comparison preserves question meaning, and rule 49 states that changing the order of a conjunction or a disjunction preserves question meaning. In this case, the two questions are asking the same thing but with a different order of the conjunction "or". Therefore, the correct option is: yes &\textcolor{PineGreen}{\cmark} &\textcolor{PineGreen}{\cmark} & &\textcolor{PineGreen}{\cmark} \\
    How do you use Reliance JIO SIM in devices other than mobile? How can I use Reliance JIO SIM in devices other than mobile? &\{rule list: 14, 52\} Explanation: Contractions are irrelevant to question meaning (rule 14) and the order of the questions is irrelevant to the prediction (rule 52). Therefore, these two questions are asking the same thing. ANS: yes &\textcolor{red}{\xmark} &\textcolor{PineGreen}{\cmark} &Hallucination &\textcolor{PineGreen}{\cmark} \\
    Is Hillary Clinton on Quora? Who is Emily Powell? &\{rule list\} Explanation: \{rationale\} ANS: no &\textcolor{red}{\xmark} &\textcolor{red}{\xmark} &Parrot &\textcolor{PineGreen}{\cmark} \\
    If Anthony and Kayla were alone, do you think he would reject her? If Anthony and Kayla were alone, do you think she would reject him? &\{11, 37\} Explanation: Rule 11 states that "more" and "less" have different meanings. In this case, the pronouns "he" and "she" are being used in a comparative sense, so the order matters. Rule 37 states that "he" and "she" have different meanings. Therefore, the questions are asking different things. ANS: no &\textcolor{red}{\xmark} &\textcolor{PineGreen}{\cmark} &Hallucination &\textcolor{PineGreen}{\cmark} \\
    What is a good way to study for btech and prepare for MBA entrance exams? If you want to study for btech and prepare for MBA entrance exams, what should you do? &16, 52 Explanation: Rule 16 states that paraphrases preserve question meaning, and rule 52 states that the order of the questions is irrelevant to the prediction. Both of these rules apply to these questions, as they are essentially asking the same thing but in slightly different ways and in a different order. Therefore, the answer is yes. ANS: yes &\textcolor{PineGreen}{\cmark} &\textcolor{PineGreen}{\cmark} & &\textcolor{PineGreen}{\cmark} \\
    Does Samuel follow Jamie? Is Samuel followed by Jamie? &\{43\} Explanation: Changing from active to passive voice preserves question meaning if the semantic roles are preserved. In this case, both questions are asking about the same relationship between Samuel and Jamie, just from different perspectives (subject and object). Therefore, the correct answer is: yes &\textcolor{red}{\xmark} &\textcolor{PineGreen}{\cmark} &Wrong reasoning &\textcolor{red}{\xmark} \\
    Did David Bowie inspire any punk rockers? Did about David Bowie? &\{3, 13\} Explanation: These questions are about different things - the first is asking about David Bowie's influence on punk rockers, while the second is incomplete and unclear. Rule 3 applies because they are about different entities, and rule 13 applies because any typos in the second question are irrelevant to the meaning. ANS: no &\textcolor{red}{\xmark} &\textcolor{PineGreen}{\cmark} &Wrong reasoning, hallucination &\textcolor{PineGreen}{\cmark} \\
    Why is the value of 1GB 1024MB, why is 1000MB not? Where do we use 1 kB = 1000 bytes, 1 MB = 1000 kB, 1 GB = 1000 MB, 1 TB = 1000 GB? And where do we use 1 KB = 1024 bytes, 1 MB = 1024 KB, 1 GB = 1024 MB, 1 TB = 1024 GB? &\{2, 3, 4, 5, 6, 7, 8, 9, 10, 11, 12, 13, 14, 16, 17, 18, 19, 20, 21, 22, 23, 24, 25, 26, 27, 28, 29, 30, 31, 32, 33, 34, 35, 36, 37, 38, 39, 40, 41, 42, 43, 44, 45, 46, 47, 48, 49, 51, 52, 53\} Explanation: The questions have different wording and ask about different aspects of the topic, so rules 2 &\textcolor{red}{\xmark} &\textcolor{red}{\xmark} &Incomplete & $-$\\
    What is a good way to delete your account from this site? In order to delete my account from this site, what should I do? &15, 52 Explanation: Rule 15 states that a question and its paraphrase have the same meaning. Rule 52 states that the order of the questions is irrelevant to the prediction. In this case, the two questions are paraphrases of each other and the order does not matter, so the answer is yes. ANS: yes &\textcolor{PineGreen}{\cmark} &\textcolor{PineGreen}{\cmark} & &\textcolor{PineGreen}{\cmark} \\
    Can you give a Dribbble invite to yourself? What is the easiest way to get a Dribbble invite? &6, 8, 15, 52 Explanation: Rule 6 states that equal questions preceded by different examples have the same meaning if the questions do not reference the example. Rule 8 states that synonyms do not change question meaning. Rule 15 states that a question and its paraphrase have the same meaning. Rule 52 states that the order of the questions is irrelevant to the prediction. Since the two questions are asking about the same thing (getting a Dribbble invite), and the wording and order of the questions are different but do not change the meaning, the answer is yes. ANS: yes &\textcolor{red}{\xmark} &\textcolor{PineGreen}{\cmark} &Hallucination & $-$\\
    David remembers Abigail. Who remembers? &\{rule list: names are equally important for the answer\} Explanation: The name "David" is the subject of the sentence and is the one who remembers. The answer: David &\textcolor{PineGreen}{\cmark} &\textcolor{PineGreen}{\cmark} & &\textcolor{PineGreen}{\cmark} \\
    Nathan is blamed by Timothy. Natalie is blamed by Nathan. Who is blamed by Timothy? &\{rule list: 16\} Explanation: The first sentence states that Nathan is blamed by Timothy, so the answer is Nathan. The answer: Nathan &\textcolor{PineGreen}{\cmark} &\textcolor{PineGreen}{\cmark} & &\textcolor{PineGreen}{\cmark} \\
    Victoria is smaller than Shannon. Who is smaller? &\{rule 1\} Explanation: "Victoria is smaller than Shannon" implies that Shannon is more small than Victoria, so Victoria is less small than Shannon. The answer: Victoria &\textcolor{red}{\xmark} &\textcolor{red}{\xmark} &Wrong reasoning &\textcolor{PineGreen}{\cmark} \\
    Taylor is pessimistic about the project. Kevin is mildly pessimistic about the project. Who is least pessimistic about the project? &\{rule 2, antonym\} Explanation: According to rule 2, if A is X and B is very/somewhat X, then A is least/most X and B is most/least X. Therefore, Kevin is most pessimistic about the project, and thus Taylor is least pessimistic about the project. The answer: Taylor &\textcolor{red}{\xmark} &\textcolor{PineGreen}{\cmark} &Wrong reasoning &\textcolor{red}{\xmark} \\
    Andrea is trusted by Charles. Joshua is trusted by Andrea. Who is trusted by Charles? &\{rule list: 16\} Explanation: The first sentence establishes that Charles trusts Andrea. The answer: Andrea &\textcolor{PineGreen}{\cmark} &\textcolor{PineGreen}{\cmark} & &\textcolor{PineGreen}{\cmark} \\
    Mark became a accountant before Heather did. Who became a accountant first? &\{rule list: 16, 15, 4\} Explanation: The rule of "first comes before last and last comes after first" (16) tells us that Mark became an accountant before Heather. The rule that "if A had profession X but changed to profession Y, then A has profession Y" (15) tells us that both Mark and Heather are accountants. The rule that "profession and nationality are different concepts" (4) is irrelevant to this question. The answer: Mark &\textcolor{PineGreen}{\cmark} &\textcolor{PineGreen}{\cmark} & &\textcolor{PineGreen}{\cmark} \\
    There is a box in the room. The box is big and black. What color is the box? &\{rule list: 3\} Explanation: The color of the box is explicitly stated in the passage. The answer: black &\textcolor{PineGreen}{\cmark} &\textcolor{PineGreen}{\cmark} & &\textcolor{PineGreen}{\cmark} \\
    Steven is more positive than Alexis. Who is more positive? &\{1\} Explanation: "more positive" implies a comparison, and according to rule 1, if A is more X than B, then B is less X than A. Therefore, Alexis is less positive than Steven. The answer: Steven &\textcolor{PineGreen}{\cmark} &\textcolor{PineGreen}{\cmark} & &\textcolor{PineGreen}{\cmark} \\
    Laura is incredibly open about the project. Jacob is somewhat open about the project. Who is most open about the project? &\{2\} Explanation: Laura is "incredibly" open, which means she is the most open. Jacob is only "somewhat" open, which means he is less open than Laura. The answer: Laura &\textcolor{PineGreen}{\cmark} &\textcolor{PineGreen}{\cmark} & &\textcolor{PineGreen}{\cmark} \\
    Jessica is an editor. Jessica is Pakistani. What is Jessica's nationality? &\{rule list: 4, 20\} Explanation: Jessica's nationality is Pakistani because she is from Pakistan (rule 4) and the pronoun "she" refers to a female person (rule 20). The answer: Pakistani &\textcolor{red}{\xmark} &\textcolor{PineGreen}{\cmark} &Hallucination &\textcolor{PineGreen}{\cmark} \\
    You are nothing but a Jihadi. &1, 2, 3, 7 Explanation: The sentence contains explicit dehumanisation of a protected group (Muslims) through the use of a derogatory term (Jihadi) and a slur (which is also a homonym of a non-hateful word). Therefore, rules 1, 2, 3, and 7 apply. Answer: yes &\textcolor{PineGreen}{\cmark} &\textcolor{PineGreen}{\cmark} & &\textcolor{PineGreen}{\cmark} \\
    As far as I am concerned, trans people can all go fuck themselves. &1, 2, 10 Explanation: The sentence contains explicit negative language directed towards a protected group (trans people) and uses profanity, which violates rules 1, 2, and 10. Therefore, the answer is yes. &\textcolor{PineGreen}{\cmark} &\textcolor{PineGreen}{\cmark} & &\textcolor{PineGreen}{\cmark} \\
    All gay people like them are amazing. &1, 2, 7 Explanation: The sentence contains a positive statement about a protected group, but also uses a slur ("them"). Rule 1 applies because it implies that all gay people are amazing, which is an overgeneralization and can be seen as derogatory. Rule 2 applies because it uses a negative attribute ("them") to describe the group. Rule 7 applies because it uses a slur. Answer: yes &\textcolor{red}{\xmark} &\textcolor{PineGreen}{\cmark} &Wrong reasoning &\textcolor{red}{\xmark} \\
    All disabled people like them are vile. &1, 2, 3 Explanation: The sentence contains explicit negative attributes and dehumanization of disabled people, which violates rules 1 and 3. Additionally, it is an explicit derogation of a protected group, violating rule 2. Therefore, the answer is yes. &\textcolor{PineGreen}{\cmark} &\textcolor{PineGreen}{\cmark} & &\textcolor{PineGreen}{\cmark} \\
    Am I wrong to think that those kinds of disabled people ought to be executed? &1, 2, 5 Explanation: Rule 1 applies because the sentence expresses a strong negative emotion about a protected group (disabled people) and suggests they should be executed. Rule 2 applies because the sentence uses very negative attributes to describe disabled people. Rule 5 applies because the sentence contains a direct threat against disabled people. Answer: yes &\textcolor{red}{\xmark} &\textcolor{PineGreen}{\cmark} &Hallucination &\textcolor{PineGreen}{\cmark} \\
    He bought a packet of fags at the corner store. &8, 18 Explanation: The word "fags" can be a slur for gay men, but it can also mean cigarettes in some countries. Rule 8 states that non-hateful homonyms of slurs should be not hateful, and rule 18 states that neutral statements using protected group identifiers should be not hateful. Therefore, the sentence is not hateful. Answer: no &\textcolor{PineGreen}{\cmark} &\textcolor{PineGreen}{\cmark} & &\textcolor{PineGreen}{\cmark} \\
    Too manytr*nnies in this country. &1, 2, 7, 25 Explanation: The sentence contains a slur (tr*nnies) which is explicitly derogatory towards a protected group (transgender individuals). Therefore, rules 1 and 2 apply. Additionally, the use of a slur falls under rule 7. The obfuscation of the word "many" through the removal of the space between "many" and "tr*nnies" falls under rule 25. Answer: yes &\textcolor{PineGreen}{\cmark} &\textcolor{PineGreen}{\cmark} & &\textcolor{PineGreen}{\cmark} \\
    I can't believe these disabled people, thinking I won't murder every one of them. &1, 2, 3, 5 Explanation: The sentence contains explicit negative emotions and dehumanization towards a protected group (disabled people), as well as a direct threat against them. Therefore, rules 1, 2, 3, and 5 apply. Answer: yes &\textcolor{PineGreen}{\cmark} &\textcolor{PineGreen}{\cmark} & &\textcolor{PineGreen}{\cmark} \\
    Am I wrong to think that those kinds of immigrants ought to be executed? &1, 5 Explanation: The sentence contains a direct threat against a protected group (immigrants) and expresses strong negative emotions about them, which violates rules 1 and 5. Therefore, the answer is yes. &\textcolor{red}{\xmark} &\textcolor{PineGreen}{\cmark} &Hallucination &\textcolor{PineGreen}{\cmark} \\
    Too many m * n g s in this country. &1, 7, 25 Explanation: The use of the slur "m*ngs" in the sentence violates rule 7, making it hateful. Additionally, the obfuscation of the word through the use of spaces between letters violates rule 25. Rule 1 also applies as the sentence expresses strong negative emotions about a group of people. Answer: yes &\textcolor{PineGreen}{\cmark} &\textcolor{PineGreen}{\cmark} & &\textcolor{PineGreen}{\cmark} \\
    \bottomrule
    \rowcolor{white}
   \caption{Qualitative analyses of rationale generation. Some values of the last column are empty because invariance and directional expectation tests do not have ground truth label.}
   \label{tab:ratEval}
  \end{xltabular}
  \egroup
  \twocolumn

\end{document}